# From Single Chatbots to Governed Agent Ecosystems: An Agentic AI Pattern Catalogue and Orchestration Framework for Mission-Critical Hospital Information Management Systems

Manideep Dhar[1*]; Ritwik Singh[2]; Sharat Chandra Kumar Manikonda[3]

[1]Chief Operating Officer and Head of Research and Development, Instil-it, Hyderabad, Telangana.
[2]Chief AI Architect Engineer, Instil-it, Hyderabad, Telangana
[3]Chief Executive Officer, Managing Director and Chief Data Scientist, Instil-it, Hyderabad, Telangana

[1]ORCID: 0009-0000-3394-2158

Corresponding Author: Manideep Dhar[1*]

**Abstract: Hospitals are racing to embed Artificial Intelligence (AI), while coping with the surge in adaptation of the technology in other industries, into the triage management, documentation, scheduling, and revenue-cycle workflows, yet most deployments remain as fragmented pilots that stall at the edge of production, exposing patients and institutions to operational fragility, ungoverned risk, and mounting technical debt. At the same time, the global AI-in-healthcare market is projected to exceed nearly USD 1 trillion by 2034, according to the report of Fortune Business Insights, amplifying the financial consequences of architectural missteps and failed scaling strategies. This research proposes a compliance-first Agentic AI pattern catalogue and orchestration framework, purposely built for Hospital Information Management Systems (HIMS), moving beyond the single Large Language Model (LLM) chatbots and towards a governed ecosystem of autonomous and semi-autonomous agents. The framework extends by adding (i) a taxonomy of Agentic roles (conversational, orchestration, reconciliation, auditing, and decision-support agents), (ii) a formal risk-stratification model that maps each pattern to risk tiers, human-in-the-loop checkpoints, and governance hooks, and (iii) a unified orchestration runtime capable of coordinating multi-agent workflows across EHR/HIMS landscapes such as Epic, Cerner, and MEDITECH. Technically, the framework combines vLLM (Virtual Large Language Model)-based inference, optimized paging memory, confidential computing, and Model Context Protocol (MCP) based on-premise deployment, enforcing end-to-end encryption and policy-as-code controls aligned with HIPAA, GDPR, the EU AI Act, India's DPDP and DISHA Acts, ISO 27001, ISO 27002, ISO 14971, and IEC 62304. Using synthetic but structurally realistic and reflecting the complexities of the hospital data generated using Synthea, and controlled pilot deployment and functional run, we exhibit how the proposed architecture is capable and efficient to reduce the documentation time, integration effort, and AI pilot attrition while constriction the governance and auditability, offering hospital leaders and governing authorities an urgently needed blueprint to convert AI investment into sustainable clinical, operational, and financial ROI.**

# I. INTRODUCTION

➢ *Problem Context and Industrial Urgency*

Over the last decade, AI in healthcare has shifted from research curiosity to strategic infrastructure, with market analyses projecting that global AI-in-healthcare spending will grow from roughly USD 29–39 billion in 2024–2025 to more than USD 500–1,000 billion by 2032–2034, as per the reports of Fortune Business Insights [1][3][Figure 1]. Recent market analysis report published by Keneally et al. (2026) [2] on behalf of Boston Consulting Group (BCG) reveals India topping the world charts with AI adoption in healthcare, with 85% successful adoption [Figure 2] [2]. This explosive growth is driven by converging pressures: chronic workforce shortages, rising documentation burden, margin compression, and escalating patient expectations for responsive, digitally mediated care [3][4][5]. Yet, alongside this investment boom, multiple independent analyses converge on a sobering pattern: between 70–80% of healthcare AI pilots never scale beyond the pilot phase, and a significant share deliver zero measurable financial benefit [3][6][7]. The study of BCG also reveals some key insights, out of which it was highlighted that 62% of the survey respondents expressed concerns over the data privacy and 59% of the respondents' expressed concerns on the reliability of the AI-driven systems in healthcare [2], and further highlighted in explicit way how the market, which is currently driven by single LLM based AI -Chatbots and AI-enabled fitness wearables, is currently anticipating reliable, enhanced data privacy and protection practical Agentic AI solutions in the healthcare market [8][9].

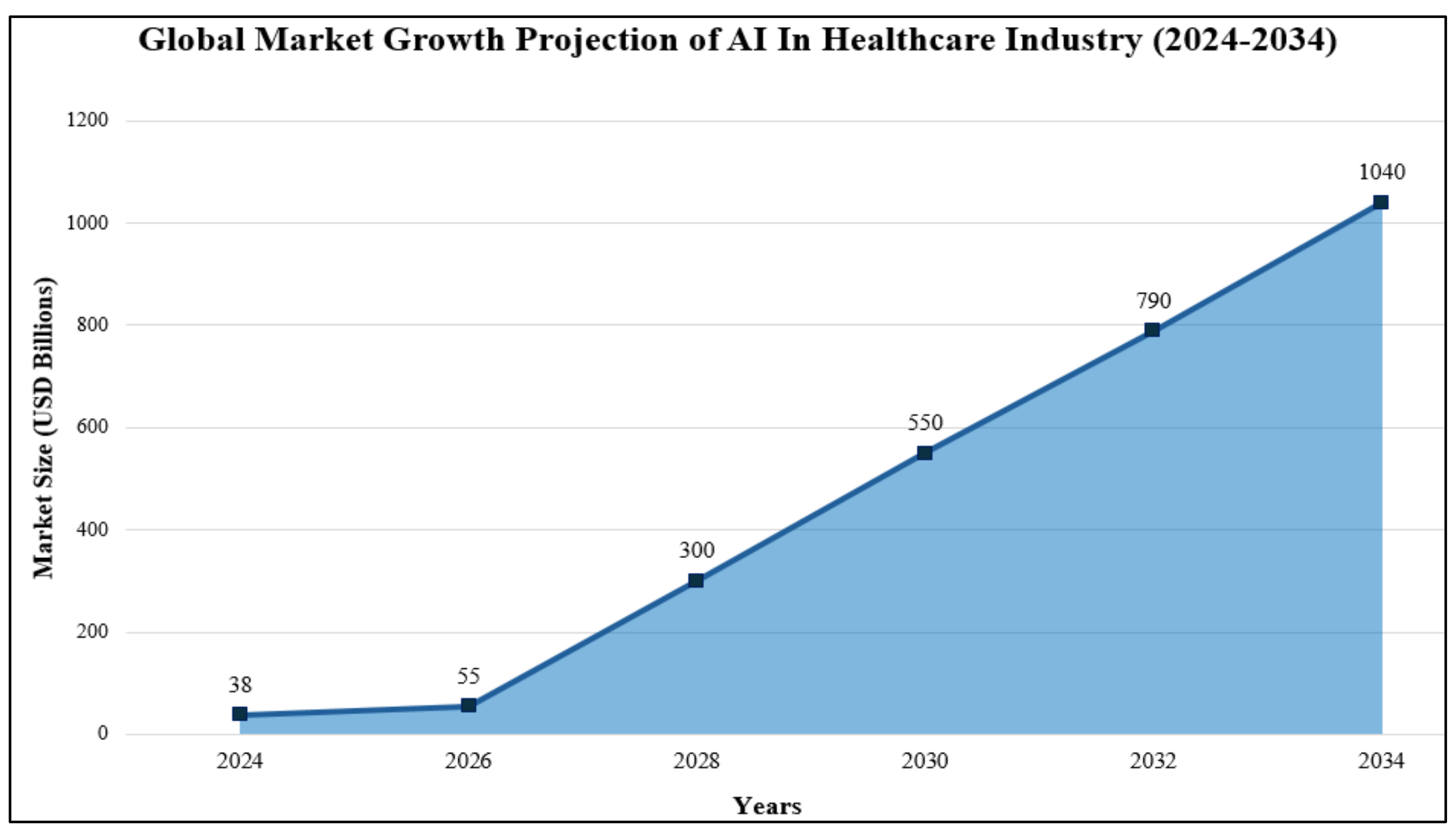

Fig 1 Market Growth Projection of AI in Healthcare Industry (2024-2034)
(Source: Fortune Business Insights) [1] [3]

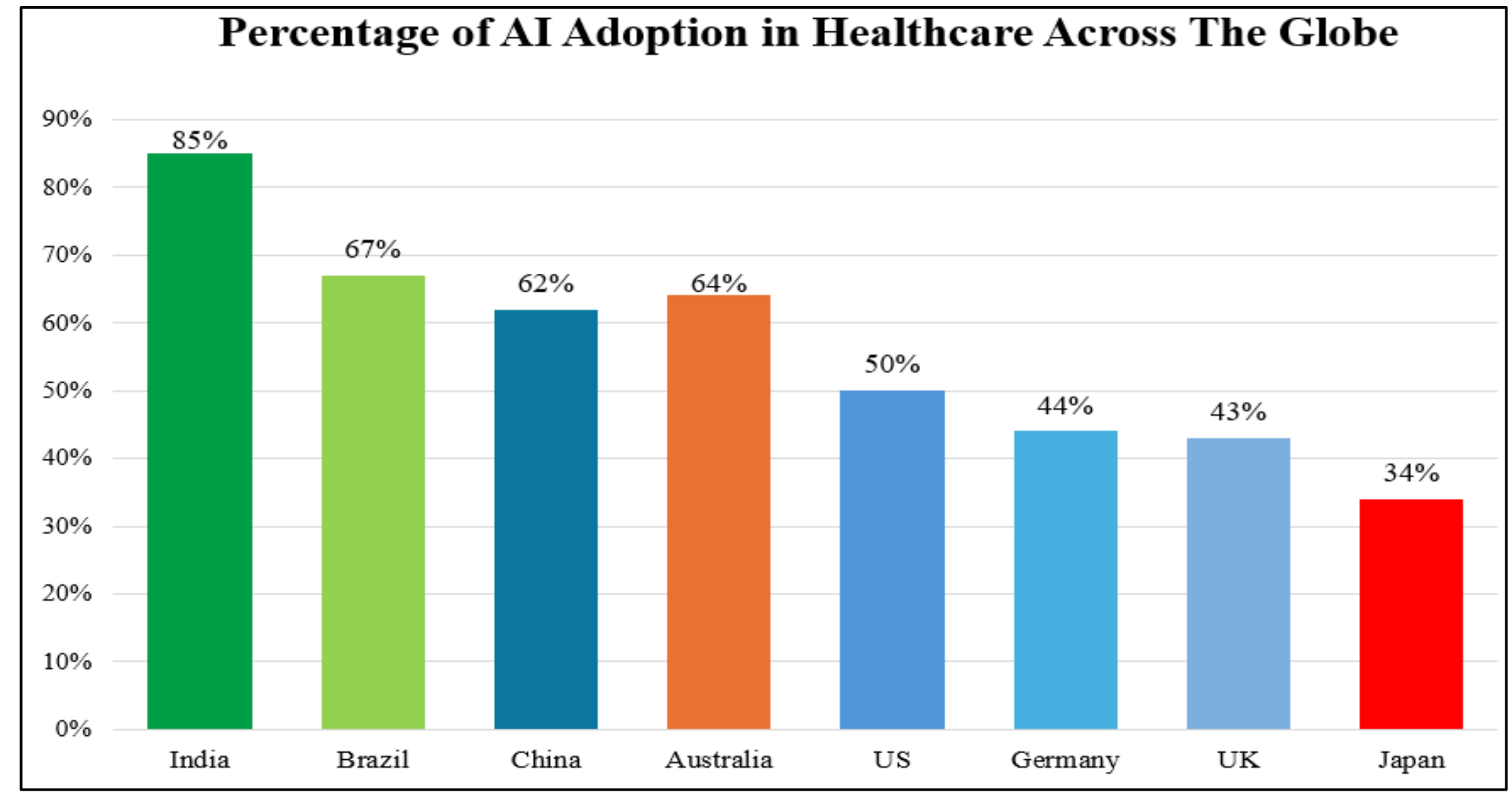

Fig 2 Percentage of AI Adoption in Healthcare Around the World.
(Source: BCG Study 2026)[2]

This pilot-to-production gap is not merely an IT nuisance; it represents a systemic risk for hospital systems that increasingly depend on AI for triage, bed management, prior authorization, and revenue-cycle integrity [3][10][11]. Each failed pilot project leaves behind orphaned integrations, redundant data pipelines, and unmaintained models that accumulate as technical and governance debt, eroding trust among clinicians and boards [12]. At the same time, clinicians in major health systems report spending around 13.5 hours per week on clinical documentation, as per the reports of NHS Survey [13] and recent published studies [3][14][15], more than a third of their working time and roughly 25% more than the average reported seven years ago [Figure 3], directly reducing patient-facing capacity and fuelling clinician burnout leading to fatigue and impairing cognitive functionality of the clinicians [16].

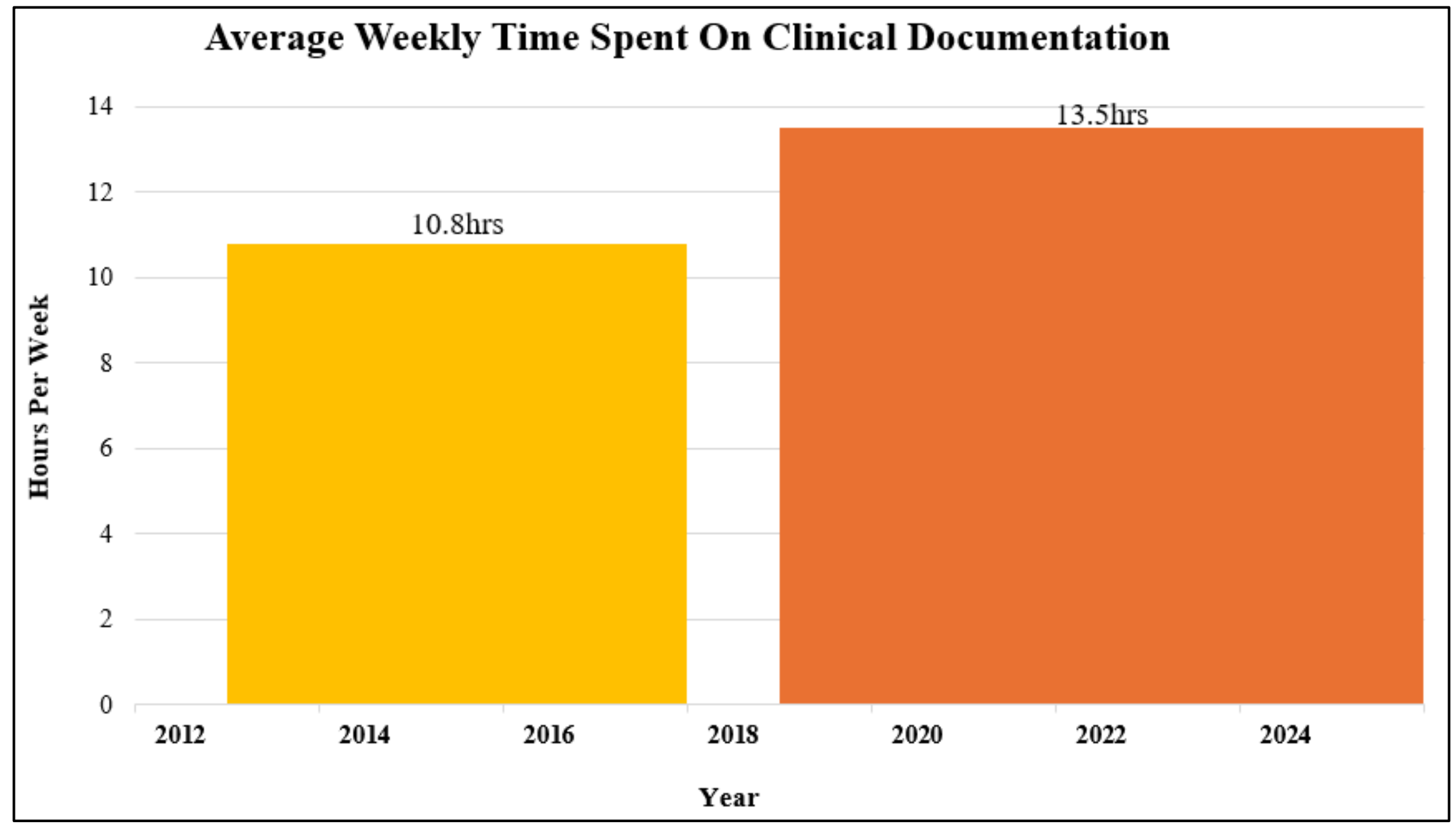

Fig 3 Graphical Representation of the Average Weekly Time Spent on Clinical Documentation Reported by the Clinicians (Source: NHS Survey) [3][13]

Without an architectural reset, restructure and governed solution, hospitals risk a future where AI adds layers of opaque complexity without materially relieving the operational and cognitive load that it was meant to alleviate.

➢ *From Single Chatbots to Governed Agent Ecosystems*

Most current "AI in HIMS" deployments take the form of single LLM chatbots or narrow machine-learning models embedded within specific applications, as examples: a billing and discharge validation agent in the finance and operations department module, an imaging classifier in the Picture Archiving and Communication System (PACS) [17], a standalone medical coding assistant, or a scheduling optimizer bolted to appointment systems [3][18][19]. These tools are typically engineered in isolation, with bespoke data feeds, proprietary agent behaviours, and ad hoc governance workflows, resulting in a patchwork of siloed algorithms rather than a coherent, governed AI platform [3][20][21].

Emerging Agentic AI approaches, multi-agent systems where specialized agents collaborate to plan, reason, and act [22][23], are particularly efficient and fit for hospitals because they mirror the multi-role nature of clinical and administrative teams. To understand this better, let's visualise how a single patient's journey legitimately involves a triage risk agent, a documentation summarization agent, a bed-allocation agent, and a billing-codes reconciliation agent, orchestrated through a shared context. However, without hospital-specific patterns, risk tiers, and orchestration rules, the Agentic systems' risk areas are becoming wider with un-auditable black boxes whose internal message passing and tool calls are invisible to compliance teams and safety committees [24][25].

What is missing today is not another chatbot but a pattern catalogue and orchestration framework that makes Agentic AI governable, auditable, and reproducible in the context of Hospital Information Management Systems. Such a framework must be explicitly aligned with regulatory regimes, HIPAA [26][27], GDPR [28][29], the EU AI Act [30][31], India's DPDP Act [32][33][34] and Digital Information Security in Healthcare Act (DISHA), India [35][36][37], and international standards such as ISO/IEC 27001 [38], ISO/IEC 27002 [39], ISO 14971 [40] and IEC 62304 [41], and must answer the hard board-level question: *"Where, exactly, could protected health information go?"*— not in general terms, but with traceable, verifiable, and enforceable design statements.

➢ *Anchoring the Compliance-First Architecture*

The companion study by Dhar et al. (2026) [3] defines a seven-layer, compliance-first architecture for hospital AI systems, including an Infrastructure and Runtime Layer, a Privacy-Preserving Data Fabric, a Model and Analytics Layer, an Agent Orchestration Layer, an Application and Workflow Layer, a Compliance and Policy Layer, and a Governance, Risk, and Assurance Layer [3]. That work demonstrates, using synthetic hospital datasets generated with Synthea [42][43][44] and the production-grade model, that such an architecture can dramatically reduces integration efforts and improve compliance coverage across HIPAA [26][27], GDPR [28][29], the EU AI Act [30][31], India's DPDP Act [32][33][34] and Digital Information Security in Healthcare Act (DISHA), India [35][36][37], and international standards such as ISO/IEC 27001 [38], ISO/IEC 27002 [39], ISO 14971 [40] and IEC 62304 [41].

This research builds directly on and remains fully aligned with the architecture demonstrated by Dhar et al. (2026) [3], including its security controls, policy-as-code framework, and privacy-preserving data fabric [3], while focusing on a narrower but crucial design problem: how to systematically specify, risk-tier, and orchestrate Agentic AI patterns embedded in mission-critical HIMS workflows. Concretely, we treat the architecture of Dhar et al. (2026) [3] as the platform foundational substrate, and we define on top of it:

- A pattern catalogue of agent roles including conversational query agents, workflow orchestration agents, reconciliation and auditing agents, and decision-support agents constrained to assistive behaviour in safety-critical contexts.
- A risk stratification and human-in-the-loop model that maps each pattern to operational risk tiers (low, medium, high, safety-critical) and prescribes governance hooks and escalation pathways.
- A unified orchestration runtime that can register, configure, and monitor multiple agents within a single HIMS-integrated control plane instead of bespoke, vendor-specific integrations.

By explicitly referencing and extending the substrate framework of Dhar et al. (2026) [3], this research ensures that architecture, security, and policy-as-code [3] are efficient enough to carry over into the proposed Agentic pattern catalogue and orchestration design.

➢ *Context of the Identified Problems*

We formalize the central research objective around the problem, where, hospitals are transitioning from single LLM chatbots to ecosystems of autonomous or semi-autonomous agents that query EHRs, schedule resources, draft documentation, and escalate alerts. However, there is no systematic taxonomy or reference implementation of safe Agentic patterns tailored to HIMS, and existing engineering practices from other industries do not directly address clinical risk, data sensitivity, and strict audit requirements under global healthcare regulations.

- *This Gap Manifests Across Several Intertwined Dimensions:*

✓ *Architectural Ambiguity:*

No shared vocabulary exists for agent roles and interaction patterns specific to HIMS (e.g., triage-to-discharge orchestration, coding reconciliation with consent logs, or risk-tiered decision-support escalation).

✓ *Governance Opacity:*

Most prototypes lack explicit mappings between agent behaviours and regulatory clauses (HIPAA Security Rule [26][27], GDPR special-category data rules [28][29][45], EU AI Act high-risk provisions [30][31][46], DPDP Act India [32][33][34] and DISHA India [35][36][37] obligations), making compliance attestation difficult and brittle.

✓ *Security and Data-Flow Uncertainty:*

Traditional cloud-native patterns often expose more services to the public internet than necessary, leaving the Fast Health Interoperability Resources (FHIR) servers, caches, and internal orchestration APIs reachable through poorly understood paths [47][48]. This contradicts the principle that "nothing that does not need the public internet should ever see the public internet at all" [49] and leaves boards unable to answer where protected health information can travel.

✓ *Infrastructure Inefficiency:*

Naive deployment of large LLMs for every interaction is incompatible with real-time hospital constraints. Without vLLM-style optimized paging and concurrency, multi-department usage would quickly drives latency spikes that are clinically unacceptable in high critical hospital operations like triage management, escalation management, or code-blue scenarios [50][51].

✓ *Deployment Fragmentation:*

Hospitals operate under diverse data-localization and sovereignty regimes; many require on-premise or sovereign-cloud deployments and cannot simply consume "AI-as-a-Service". No unified pattern exists to express the same Agentic design across on-premise Model Context Protocol (MCP) based, hybrid, and full-cloud topologies while preserving identical governance semantics [52][53][54].

✓ *The Consequence is a Growing Implementation Gap:*

While AI capabilities expand, hospitals lack a governed, reusable blueprint to move from pilots to scalable, compliant, multi-agent HIMS platforms.

➢ *Objectives and Core Innovation*

In response to the problem areas identified, this research pursues four intertwined objectives:

- *Agentic Pattern Catalogue for HIMS*

To define a taxonomy of Agentic AI patterns embedded in hospital information management systems, covering:

✓ Conversational query agents for clinicians, administrators, and patients, integrated through FHIR [55] or SMART-on-FHIR [56] and constrained by policy scopes.

✓ Workflow orchestration agents coordinating multi-step processes such as triage management, imaging process, bed assignment, discharge process, billing management.

- ✓ Reconciliation and auditing agents that continuously cross-check documentation, billing, consent, and access logs.
- ✓ Decision-support agents that remain strictly assistive in high-risk clinical contexts, with enforced human-in-the-loop review and override mechanisms.

- *Risk Stratification and Governance Mapping*

To introduce a formal risk stratification model that classifies each pattern into operational risk tiers (low, medium, high, safety-critical) and aligns them with EU AI Act [30][31][46] risk categories and other regulatory frameworks [26][27] [28][29][45][32][33][34][35][36][37]. For each tier, we define required human oversight, escalation conditions, logging obligations, and policy-as-code rules that can be enforced within the Compliance and Policy Layer defined in the published work of Dhar et al. (2026) [3].

- *Unified Orchestration Runtime and Deployment Variants*

To specify a unified orchestration runtime that allows hospitals to register, configure, and monitor multiple agents under a single HIMS-integrated control plane, rather than maintaining point-to-point integrations per vendor. The runtime is designed to support:

- ✓ MCP-based on-premise deployments, where Model Context Protocol (MCP) [54] servers serve as the integration surface to local tools and EHRs without exposing PHI to the public internet.
- ✓ Hybrid cloud variants where confidential-computing enclaves, Virtual Private Cloud (VPC) [57] peering, and VPNs [58] confine PHI to private subnets while offloading only de-identified signals or encrypted embeddings.
- ✓ Cloud-native deployments in HIPAA-eligible [26], GDPR-compliant [28] regions, still obeying the principle that only a minimal, hardened ingress is exposed externally.

- *vLLM-Optimized Inference for Hospital Operations*

To develop an inference architecture built on Virtual Large Language Models (vLLM) that uses paged attention [59], efficient Key Value cache (KV-cache) sharing [60], and workload-aware scheduling to serve multiple concurrent clinical and operational users with bounded latency. We explicitly design memory management policies to align with hospital time-critical SLAs, and we exhibit that vLLM-based orchestration can meet sub-second response targets for high-volume scenarios when appropriately sized and configured appropriately.

- *Encryption-First, Confidential Intelligence, and Address the Governing Question "Where PHI Can Go?"*

The proposed architecture adopts a zero-trust, encryption-as-default stance where any data that can be accessed is treated as potentially leakable and must therefore be protected at-rest, in-transit, and in-use. By design, movement of PHI is restricted to controlled, documented pathways, each mediated by policy-as-code rules and keyed to roles, scopes, and jurisdictions [3]. This is the foundation of the confidential intelligence system where we present a system that can operate on sensitive information without ever exposing it unnecessarily.

➢ *Positioning Within AI, HIMS, and Emerging Regulations*

Our research sits at the intersection of software architecture, AI safety, and healthcare regulation. While prior work has described generic layered architectures for healthcare AI platforms, few studies explicitly map architectural layers to regulatory clauses (e.g., HIPAA Security Rule [26][27] safeguards, GDPR Article 9 special-category data restrictions [28][75], EU AI Act Articles 9–15 [30][31][46][76], DPDP obligations [32][33][34], DISHA's [35][36][37] prohibitions on commercial reuse of health data in India, ISO/IEC 27001[38], ISO/IEC 27002 [39], ISO 14971 [40], IEC 62304 [41]). The study of Dhar et al. (2026) [3] is a notable exception, providing a compliance-first Agentic platform blueprint, this paper extends their work to agent patterns and orchestration semantics.

With that we aim to provide a compliance-aligned and practically tested blueprint rather than just another technical reference, with which hospital leaders should be able to trace, for each agent pattern and orchestration flow, which safeguards are implemented, where PHI can move, and how residual risks are managed within the institution's governance and risk appetite.

## II. LITERATURE REVIEW

➢ *AI In Healthcare and HIMS: Promises and Persistent Gaps*

A substantial body of work surveys AI applications in healthcare, highlighting opportunities in imaging, triage, diagnosis, patient monitoring, and administrative automation [77][78][79]. Jiang et al. (2017) [80] and later reviews describe how machine learning and deep learning models have achieved high performance across radiology, pathology, and risk prediction tasks, while emphasizing the need for explainability and integration into clinical workflow [81][82][83][84][85]. Longitudinal bibliometric analyses similarly document exponential growth in AI-related medical publications and underscore the central role of large language models and multimodal deep learning in recent years [86][87][88][89]. However, much of this literature treats AI models as point solutions, evaluated against specific datasets or tasks, without specifying the platform architectures and integration patterns needed to embed these models into complex Hospital Information Management Systems (HIMS). As a result, successful model performance in controlled experiments does not readily translate into reliable, governable performance across heterogeneous EHRs, PACS , Laboratory Information System (LIS), and billing systems [90]. This disconnect between model-centric research and systems-level deployment is increasingly recognized as a bottleneck to realizing AI's promised impact in hospitals [91][92].

The architecture demonstrated in the work of Dhar et al. (2026) [3] directly responds to this gap by proposing a multi-layered, compliance-first hospital AI platform that unifies infrastructure, data, models, agents, compliance, and governance under a single conceptual framework [3]. Our contribution builds on that foundation by articulating Agentic patterns and orchestration semantics that can be systematically applied across HIMS modules, thus operationalizing AI capabilities at scale rather than as isolated tools.

➢ *Agentic AI, Multi-Agent LLMs, and Clinical Safety*

Recent works in Agentic AI and multi-agent LLM systems has shown that decomposing complex tasks into collaborative agents can improve performance, robustness, and interpretability in non-clinical domains such as tool-augmented reasoning, code generation, and multi-step planning [93][94][95]. Wu et al. (2023) [96] and Dong et al. (2025) [97] demonstrated how multi-agent architectures have showcased where specialized agents (planner, executor, critic, tool-caller) interact over shared memory or message buses, often yielding more stable results than monolithic LLM prompts as task complexity grows [96][97].

In healthcare, early explorations have demonstrated the potential of multi-agent LLMs for care coordination, clinical summarization, and patient communication, but these studies typically run in sandboxed environments and do not confront the full HIMS integration, audit, and regulatory landscape [98][99][100]. Moreover, many agent frameworks assume internet-connected tools and APIs by default [101][102], a non-starter for hospital environments where PHI must remain inside VPCs, VPNs, or on-premise networks and where external packet transactions must be tightly controlled [102].

- *Our Research Diverges from Generic Agent Frameworks in Three Important Ways:*

✓ We explicitly classify agent patterns by clinical and operational risk, constraining high-risk decision-support agents to strictly assistive roles with mandatory human oversight and forbidding autonomous action in safety-critical contexts.
✓ We design orchestration and tool-calling semantics that assume no direct public internet access from critical components; instead, agents consume strictly governed, policy-gated endpoints within private subnets and, where necessary, call out through MCP servers or carefully curated integration gateways.
✓ We attach agent behaviours to policy-as-code [3] controls and audit trails, ensuring that every tool call, data access, and cross-agent message can be reconstructed and justified in governance and incident review.

In doing so, the proposed pattern catalogue bridges the gap between powerful generic Agentic AI mechanisms and the safety-critical, heavily regulated reality of hospital operations.

➢ *AI Governance, Regulation, and Policy-As-Code*

The importance of AI governance in healthcare has been documented across multiple strands of literature [3][103][104][105]. Policy analyses highlight how regulations such as HIPAA [26][27], GDPR [28][29], and EU AI Act [30][31] impose obligations around risk management, transparency, logging, data minimization, and human oversight, especially for high-risk AI systems that influence diagnosis or treatment. India's proposed DISHA [35][36][37] and enacted DPDP Act [32][33][34] extend these obligations by emphasizing patient ownership of health data, prohibiting commercial reuse, and mandating consent, purpose limitation, and data localization.

Reddy et al. (2020) [106] and subsequent works argue that governance maturity, not just model accuracy, determines whether healthcare AI can move from pilot to sustainable production use, emphasizing risk-tiered oversight, lifecycle monitoring, and multi-stakeholder governance structures [106]. NIST's AI Risk Management Framework [107] similarly advocates risk-proportionate controls, recognizing that different AI systems warrant different levels of scrutiny based on their impact and potential harms [107]. However, much of the governance literature remains organizational and policy-level, providing frameworks and checklists rather than executable architectures.

The research work of Dhar et al. (2026) [3] takes a decisive step towards operationalizing governance through a Compliance and Policy Layer [Fig] that encodes regulatory obligations as machine-enforceable rules applied to data, models, and agents [3]. Its functional Policy Engine enforces role-based access control (RBAC) [70], jurisdiction-specific consent and data-transfer rules, and necessary constraints for billing vs. clinical roles, illustrating how HIPAA [26][27], GDPR [28][29], EU AI Act [30][31], DPDP [32][33][34], and DISHA [35][36][37] requirements and regulations can be translated and enforced as runtime decisions rather than static documentation. This research extends the concept by binding Agentic patterns and orchestration flows to explicit risk tiers and policy hooks, turning governance into a first-class design dimension for agent systems embedded in HIMS.

➢ *Documentation Burden and Workflow Automation*

A parallel stream of literature examines the administrative and documentation burden in healthcare and its implications for burnout, patient safety, and productivity [108][109][110]. Surveys across the NHS [111] and other systems show clinicians spending approximately 13.5 hours per week on documentation with a 25% increase over seven years [3][111]. Additional analyses report that for every hour of direct patient care, clinicians may spend nearly two hours on documentation or administrative tasks, with substantial overtime work dedicated to catching up on notes [112][113][114].

These studies consistently identify documentation as a high-impact target for automation, but they also caution that naive automation can introduce new risks: copy-and-paste errors, propagation of hallucinated content, and loss of critical nuance in clinical narratives [115][116]. Machine Learning (ML) and LLM-based documentation assistants have demonstrated potential to reduce documentation time and cognitive load, but evaluations often focus on user satisfaction and time saved, not on integration into governance frameworks or effect on medico-legal risk [117][118][119].

By embedding conversational query and documentation agents within a governed Agentic architecture, our proposed framework aims to translate time savings into credible, auditable ROI by reductions in manual effort, fewer SLA breaches, improved throughput, and reduced rework, all traced back to specific agent patterns and orchestration flows.

➢ *Hospital AI Platforms and Multi-Layer Architectures*

Several recent works have proposed multi-layer architectures for hospital AI platforms, often converging on a set of layers spanning infrastructure, data, models, applications, and security/compliance [120][121]. These frameworks correctly emphasize the need for standardized data access, model registries, and monitoring, but they often remain too generic to capture real HIMS complexity, especially the heterogeneity of EHRs, PACS [17], LIS, and billing systems, and the intricate interplay between clinical and financial workflows [121][122].

The work of Dhar et al. (2026) [3] extends this line of research by explicitly adding an Agent Orchestration Layer, a Compliance and Policy Layer, and a Privacy-Preserving Data Fabric [3], and by validating the architecture that demonstrated reduced integration effort, improved compliance coverage, and cross-jurisdictional deployment readiness across on-premise, hybrid, and cloud deployment and production configurations [3]. Their results show, for example, that platform-based deployment can reduce integration effort for new tools by approximately 70–80% relative to siloed point solutions, primarily by reusing shared data fabric and policy infrastructure [3].

- *Our Current Study Specializes and Deepens this Architecture Along Two Axes:*

✓ Within the Agent Orchestration Layer, we define concrete Agentic AI patterns, their inputs and outputs, interaction topologies, and risk-tiered constraints.

✓ Across the Compliance and Policy and Governance layers, we map these patterns to regulatory obligations and risk controls, yielding an actionable agent pattern catalogue that can be directly encoded in policy-as-code engines and deployment templates.

In that sense, the present work transforms the architecture proposed by Dhar et al. (2026) [3] into a pattern-driven orchestration framework for mission-critical HIMS, targeting scalable, governed agent ecosystems rather than isolated automated tasks.

➢ *Literature Gap Analysis in Current Platforms and the Published Researches*

Exploring across the literature, several gaps and loopholes become evident:

- Lack of HIMS-specific agent patterns: Existing Agentic AI research seldom distinguishes between low-risk administrative agents and high-risk clinical decision-support agents in terms of oversight, escalation, and audit obligations.
- Insufficient alignment between architecture and regulation: Even advanced platform proposals often reference HIPAA [26][27], GDPR [28][29], EU AI Act [30][31], DPDP Act India [32][33][34], or DISHA India [35][36][37] requirements and regulations at a high level without binding architectural components to concrete clauses or risk controls.
- Under-specified deployment models for data localization: Few studies detail how the same AI architecture can operate across on-premise, hybrid, and cloud settings while preserving governance semantics and data localization constraints.
- Missing integration of confidential computing and zero-trust networking: While the security community has advanced confidential computing, Trusted Execution Environments (TEEs) [68], and zero-trust principles, these are rarely woven into HIMS AI architectures as first-class design requirements.
- Limited exploration of vLLM and memory optimization for hospital workloads: Most performance studies of LLM infrastructure are generic; they do not analyse the unique concurrency, latency, and safety requirements of multi-department hospital operations.

➢ *Our Study Closes These Gaps by:*

- Providing a pattern catalogue tailored to hospital information management, with explicit risk tiers and interoperability patterns.
- Embedding policy-as-a-code governance from the study of Dhar et al. (2026) [3] into each pattern, aligning it with HIPAA [26][27], GDPR [28][29], EU AI Act [30][31], DPDP [32][33][34], DISHA India [35][36][37], and relevant ISO standards [38][39][40][41].
- Defining deployment blueprints for MCP-based on-premise [54], hybrid, and full-cloud configurations with confidential computing, VPC segmentation [57], and strict ingress controls.
- Designing and evaluating a vLLM-based inference architecture [59] tuned for multi-user, multi-department hospital workloads, optimizing paging memory and concurrency to meet clinical time constraints.

In doing so, this research lays the conceptual foundation for a series of succession studies: each concrete components (e.g., Emergency Department (ED) triage, operating room scheduling, facility energy optimization, or claims denials management) can be framed as a specific composition of patterns from this catalogue, instantiated within the study of Dhar et al. (2026) [3] and their aligned architecture to deliver governed, globally compliant, Agentic AI for HIMS.

## III. SCOPES AND NOVELTY OF THE RESEARCH

➢ *Scopes of the Enhanced Architecture*

The proposed architecture is deliberately scoped as an extension of the seven-layer, compliance-first Agentic AI platform defined in the published study of Dhar et al. (2026) [3], with a focus on mission-critical Hospital Information Management Systems (HIMS) rather than generic enterprise AI or standalone large language model tools. It is designed to operate as a platform capability spanning clinical, operational, and financial workflows, while keeping the foundation layer stack and responsibilities as defined in the work of Dhar et al. (2026) [3] intact.

Within this seven-layered framework [3], the scope of the current work is to enrich the Infrastructure and Execution Runtime Layer, Privacy-Preserving Data Fabric Layer, Model and Analytics Layer, Agent Orchestration Layer, and Compliance and Policy Layer [3][Fig] so that hospitals can systematically deploy Agentic AI in healthcare across real HIMS environments such as the "Big Three in Acute Care"- Epic, Cerner, and MEDITECH [123]. Concretely, the extended architecture supports end-to-end workflows including emergency triage, diagnostic imaging coordination, bed and theatre scheduling, discharge management, clinical documentation, coding, claims, reconciliation, and access-log auditing, framing each of these as a composition of conversational, workflow orchestration, reconciliation, auditing, and decision-support agents attached to specific risk tiers. The design assumes heterogeneous deployment topologies: on-premise in highly regulated or data-localisation-sensitive jurisdictions, hybrid cloud where confidential computing enclaves and VPC peering [57] to protect the PHI, and HIPAA [26][27] eligible or GDPR [28][29]compliant cloud-native configurations, while preserving the same policy-as-code semantics [3] across all variants.

The Privacy-Preserving Data Fabric Layer [Fig] remains the single access surface to clinical and operational data [3], but the current research extends its scope to expose agent-aware, encounter-scoped views for example, triage encounters with risk tiers and severity scores, workflow event streams annotated with Agentic roles and governance hooks, without altering any patient_id or baseline clinical attributes inherited from the dataset used in the research work of Dhar et al. (2026) [3]. These arrangements are engineered explicitly for the Agentic AI system, where they provide the minimal sufficient context for the vLLM-backed conversational agents, triage decision-support models, reconciliation agents, and auditing agents to operate safely under the regulations of HIPAA [26][27], GDPR [28][29], the EU AI Act [30][31], India's DPDP [32][33][34] and DISHA Acts [35][36][37], ISO/IEC 27001 [38], ISO/IEC 27002 [39], ISO 14971 [40], and IEC 62304 [41]. The scope intentionally excludes non-hospital domains (e.g., outpatient telehealth platforms not integrated with HIMS) and non-regulated consumer-facing AI, focusing instead on high-risk, regulated hospital AI use cases in which Agentic AI orchestration, confidential computing, and encryption-first design materially influence patient safety and compliance.

At the orchestration level [3][Fig], the extended architecture is scoped to support the unified, pattern-driven control over multiple agents and models of the foundational architecture [3] instead of prescribing specific vendor products. A central unified orchestration runtime is introduced within the Agent Orchestration Layer to register, configure, and observe multiple agents under a single HIMS-integrated control plane, using Model Context Protocol (MCP) servers [54] as on-premise integration surfaces for tools and EHR APIs in scenarios where PHI must never leave private subnets. This runtime is intentionally designed to be LLM-agnostic but vLLM [59] optimised so that the hospitals can swap and access the underlying large language models in real-time (multiple users at a time) as long as they comply with the paging, KV-cache [60], and latency guarantees encoded in the runtime, and as long as they operate within the zero-trust, encryption-as-default posture of the Infrastructure and Runtime Layer [3][Figure 4].

Finally, the scope of the platform includes quantitative evaluation and governance-ready observability but does not claim to cover every possible future modality or algorithm. The architecture provides slots for multi-modal models (imaging, notes, tabular) and privacy-preserving machine learning (federated learning, differential privacy) exactly as outlined in the study of Dhar et al. (2026) [3], and this research extends those slots with Agentic pattern-level metrics (task completion reliability, escalation frequency, time-to-resolution, override rates) and risk-tier distributions that can be surfaced to risk committees and regulators. In this way, the scope remains tightly focused on mission-critical hospital AI systems while providing enough generality to accommodate new agent patterns and regulatory updates without re-architecting the platform.

➢ *Novelty and Contribution*

The novelty of the current research arises from its innovative role as a pattern-level, risk-aware extension of the foundational architecture originally developed by Dhar et al. (2026) [3], turning a hospital-specific, compliance-first AI platform into a governed ecosystem of Agentic AI patterns tailored to Hospital Information Management Systems (HIMS). Where the published research of Dhar et al. (2026) [3] and their foundational Multi-Agent AI Architecture introduced the seven-layer structure [Fig], privacy-preserving data fabric, and initial multi-agent orchestration for triage and workflow optimisation [Fig][3], the current research study contributes with a taxonomy of Agentic roles, a formal risk-stratification model, and a unified orchestration runtime that together transform the Agentic AI from an experimental add-on into a first-class, policy-governed platform capability in HIMS.

First, this research offers a hospital-specific Agentic Pattern Catalogue that explicitly defines five core classes of Agentic AI in healthcare: Conversational Query Agents, Workflow Orchestration Agents, Reconciliation Agents, Auditing Agents, and Decision-Support Agents, and embeds these roles in both the data model (via Agentic_role, agent_pattern, and governance hooks) and the orchestration runtime. Existing multi-agent LLM frameworks rarely provide such HIMS-aware semantics; they typically define generic planner/worker/critic roles without accounting for the distinctions between clinical decision-support, coding reconciliation, or access-log auditing under HIPAA [26][27] and EU AI Act [28][29] obligations. By binding each pattern to specific HIMS workflows (triage, imaging, bed assignment, discharge, billing) and to the foundational Agent Orchestration Layer designed by Dhar et al. (2026) [3], this research closes a critical gap between generic multi-agent literature and the concrete realities of hospital information systems.

Second, the research introduces a formal risk-stratification model for Agentic patterns, mapping each agent and workflow step to operational risk tiers: low, medium, high, safety-critical and aligning these tiers with EU AI Act [28][29] risk categories and other regulatory regimes as furnished and mapped by Dhar et al. (2026) [3]. Unlike prior work that treats governance as a static checklist, the proposed model treats risk tier as a runtime control variable where it influences whether a given agent invocation is permitted, whether human-in-the-loop review is mandatory, how deep logging and explanation must be, and which cryptographic and network controls (e.g., confidential computing, VPC isolation [57], MCP-only integration [54]) must be enforced before PHI can be accessed. This yields a regulation-aligned, policy-as-code-driven mapping from agent patterns to concrete obligations under HIPAA's [26][27] minimum necessary standard, GDPR Article 9 [28][75], EU AI Act Articles 9–15 [30][31][46][76], DPDP [32][33][34] and DISHA [35][36][37] data localisation and commercial-reuse restrictions, and ISO 27001 [38], ISO 27002 [39], ISO 14971 [40], and IEC 62304 [41] control sets [3].

Third, the research contributes with integration of: (a) a unified orchestration runtime that natively integrates vLLM-based inference, (b) MCP-based on-premise deployment, and (c) confidential computing within the existing seven-layered foundational architecture of Dhar et al. (2026) [3], rather than treating these as external add-ons.

This runtime operationalises three non-obvious design innovations: (i) Agentic AI must be engineered to work without direct public internet access for critical components, relying instead on MCP servers in private subnets to present tools and HIMS APIs, (ii) vLLM's paged attention and KV-cache sharing are not merely performance tweaks but clinical safety enablers, allowing the platform to meet sub-second response targets for multi-departmental loads when triage and escalation decisions are time-critical; and (iii) confidential computing must be treated as part of the runtime, not an optional security add-on, so that at-rest, in-transit, and in-use PHI protection genuinely answers the board-level question "*Where exactly can protected health information go?*" with precise, testable statements.

Fourth, this study provides a data-driven, implementation-grade validation of the Agentic extension, using the same triage and workflow datasets which were derived with Synthea [42] and used in the foundational study of Dhar et al. (2026) [3], extended without changing any patient IDs or base records, and a production-ready Python implementation that reuses the foundational feature engineering and triage model logic [3] while layering risk-tier and Agentic semantics on top. Unlike other conceptual architectures, the proposed design is backed by concrete CSV datasets, model metrics, and orchestration outputs that can be executed end-to-end in a sandbox and then lifted into hospital environments as production deployment. This makes the research not only a theoretical contribution but also a deployment-ready architecture for CIOs, CMIOs, and Chief Data Officers seeking HIPAA [26][27]-compliant AI, GDPR [28][29]-conformant AI, and EU AI Act [28][29] -ready high-risk system designs in a single HIMS-centric framework.

This transforms individual "AI solutions" into reusable, composable agent patterns governed by a common control plane, making the platform inherently more scalable, auditable, and globally acceptable than ad hoc chatbot deployments or isolated machine-learning models.

## IV. METHODS AND METHODOLOGY

➢ *Research Methodology*

The present work is explicitly positioned as an enhancement of the seven-layered compliance-first hospital AI architecture [Figure 4] designed and defined in the published work of Dhar et al. (2026) [3] and not as a replacement for it. The foundational work of Dhar et al. (2026) [3] proposed a hospital-specific platform that couples AI capabilities to concrete regulatory, governance, and workflow requirements [3]. Our current research builds directly on that framework by enriching the several layers of the foundational study of Dhar et al. (2026) [3] with a taxonomy of Agentic roles, a formal risk-stratification model, and a unified orchestration runtime that incorporates vLLM [59] and MCP-based [54] deployments.

For chronological and conceptual context, we would first recall the seven layers as defined in the foundational architecture of Dhar et al. (2026) [3] [Figure 4]

- *Infrastructure and Execution Runtime Layer*

This layer established the foundational computing environment by providing the essential compute, storage, and networking resources required for system operations, while optionally incorporating secure enclave technologies [3]. It was designed to support deployment across on-premises infrastructures, private cloud environments, as well as HIPAA [26][27] eligible and GDPR [28][29] compliant public cloud platforms [Figure 4].

- *Privacy-Preserving Data Fabric Layer*

This layer provided standardized, policy-driven access to Hospital Information Management Systems and Electronic Health Records (HIMS/EHR), Laboratory Information

Systems (LIS), Picture Archiving and Communication Systems (PACS), and financial platforms [3]. It incorporated de-identification mechanisms, data quality pipelines, federated learning orchestration endpoints, and secure enclave connectors to facilitate governed and privacy-conscious data exchange across systems [3] [Figure 4].

- *Model and Analytics Layer*

This layer provided reusable model registries and services for the training, evaluation, and monitoring of classical machine learning, deep learning, and large language model (LLM)-based components [3]. It maintained metadata pertaining to data provenance, model performance, and associated risk characteristics to support governance, traceability, and lifecycle management [3] [Figure 4].

- *Agent Orchestration Layer*

This layer provided a multi-agent execution environment in which orchestrators, domain-specific agents, and safety or critic agents interacted through shared schemas and toolsets [3]. It facilitated coordinated operations across HIMS integrations, knowledge retrieval systems, and computational utilities to support complex, context-aware workflows [3] [Figure 4]

- *Application and Workflow Layer*

This layer encompassed clinical, operational, and financial applications that embedded Agentic workflows within EHR [157] front ends, clinician worklists, imaging workstations, and revenue cycle management dashboards [3] [Figure 4] Where feasible, it leveraged FHIR and SMART on FHIR standards to promote interoperability and seamless integration across healthcare systems [3].

- *Compliance and Policy Layer*

This layer established a centralized governance and policy-as-code framework that translated regulatory requirements [3], including HIPAA [26][27], GDPR [28][29], the EU AI Act [30][31], DISHA India [35][36][37], DPDP Act [32][33][34], ISO/IEC 27001 [38], ISO/IEC 27002 [39], ISO 14971 [40], and IEC 62304 [41], along with institutional policies, into enforceable controls governing data access, tool utilization, logging, and lifecycle management activities [3][Fig].

- *Governance, Risk, and Assurance Layer*

This layer provided cross-cutting capabilities for AI risk management, model inventory oversight, incident response, post-deployment monitoring, audit preparedness, and stakeholder reporting [3]. It was aligned with ISO 42001 [124] and broader enterprise risk management frameworks to support continuous assurance, accountability, and organizational governance [3] [Figure 4]

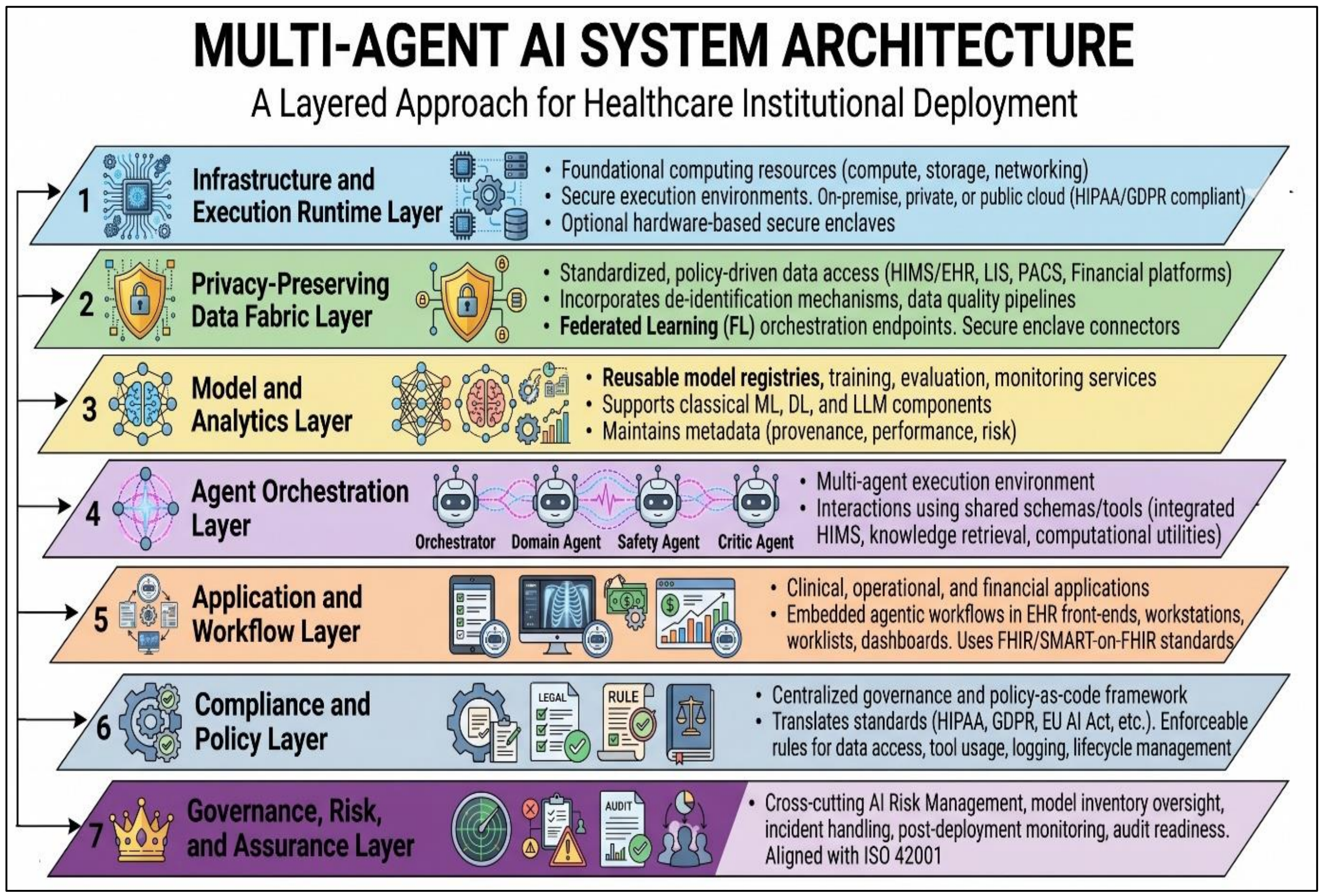

Fig 4 The Foundational Architecture of Multi-Agent Hospital AI System, Designed by Dhar et al. (2026) [3] in the Foundational Study.

The foundational study of Dhar et al. (2026) [3] treated Compliance and Policy Layer and Governance, Risk, and Assurance Layer as the control planes that determined how all other layers were configured and observed over time. The current study preserves that design decision and uses these layers as the primary anchors for adding Agentic risk tiers and governance hooks.

- *The Present Research Does not Replace that Architecture, Instead it Enhances it Along Four Axes:*

- ✓ In the Infrastructure and Execution Runtime Layer [3], we extend it by introducing standardized MCP-based [54] on-premise integrations and confidential-computing enclaves, enabling trusted system interoperability and hardware-protected execution of sensitive hospital AI workloads without altering the underlying compliant compute, storage, and networking infrastructure. Together, these extensions transform the runtime from a secure infrastructure into a trusted, interoperable execution platform capable of supporting privacy-critical, enterprise-grade multi-agent hospital information management AI ecosystem [Fig][Fig].
- ✓ The extended architecture augments the foundational Privacy-Preserving Data Fabric layer with confidential-computing enclave connectors, enabling secure data exchange into trusted execution environments [68]. This extends data protection beyond secure access and transmission to include hardware-protected processing, ensuring end-to-end confidentiality for AI-driven healthcare workflows [Fig][Fig].
- ✓ Within the Agent Orchestration Layer [3], we introduce a formal taxonomy of Agentic AI roles: conversational query agents, workflow orchestration agents, reconciliation agents, auditing agents, and decision-support agents, and encode these roles in the post-processed datasets via derived attributes: Agentic_role and agent_pattern and in the unified orchestration runtime as concrete Python agent classes as ConversationalQueryAgent, WorkflowOrchestrationAgent, ReconciliationAgent, AuditingAgent, TriageDecisionSupportAgent. We also add a formal risk stratification scheme where triage encounters get a severity_score and risk tier: low, medium, high, safety_critical, derived from age, co morbidities, prior utilisation, and vitals, along with the high_risk label [Fig] [Fig] derived from the work of Dhar et al. (2026) [3].

- ▪ Conversational Query Agents: These agents provide conversational question answering and contextual explanations for clinicians, administrators, and patients. Integrated with HIMS through FHIR [55] and SMART-on-FHIR [56], they leverage vLLM [59] to deliver context-aware, role-appropriate interactions while adhering to organisational policies and access controls [Fig][Fig].
- ▪ Workflow Orchestration Agents: These agents coordinate and manage end-to-end clinical and administrative workflows by orchestrating sequential tasks, such as triage, imaging, bed assignment, discharge, and billing, ensuring efficient, policy-compliant execution across interconnected hospital processes [Fig][Fig].
- ▪ Reconciliation Agents: These agents validate and reconcile clinical and financial records by cross-checking documentation, coding, charges, and claims, while identifying inconsistencies, anomalies, or potential discrepancies to improve data integrity and operational accuracy [Fig][Fig].
- ▪ Auditing Agents: These agents continuously monitor access logs, service-level agreement (SLA) compliance, and workflow activities to identify anomalies, support governance, and facilitate both internal oversight and regulatory audit requirements [Fig][Fig].
- ▪ Decision-Support Agents: These agents provide evidence-informed triage and treatment recommendations to support clinical decision-making while remaining strictly assistive rather than autonomous, particularly for high-risk and safety-critical scenarios where human oversight is mandatory [Fig][Fig].
- ▪ Workflow tasks get a risk_tier based on task_type, department, and SLA breach signals (e.g., triage/medication in ED/ICU are flagged as high/safety critical while coding, claims and accounting are medium). Each risk tier is then tied to governance semantics:
- ▪ High / safety-critical → mandatory human oversight (oversight_required = 1) and stricter policy checks [Fig][Fig].
- ▪ Unified orchestration runtime: The extended Agent Orchestration Layer [3] is further anchored by a Unified Agent Orchestrator [Fig][Fig], which coordinates agent execution across the platform while maintaining policy-driven governance. Its core capabilities include:
- ▪ Accessing unified triage and workflow datasets through the DataFabricView [Fig][Fig], a lightweight abstraction over the Privacy-Preserving Data Fabric layer of the foundational architecture Dhar et al. (2026) [3][Fig].
- ▪ Loading the extended triage risk model from the Model and Analytics Layer [3][Fig] to enable context-aware and risk-informed orchestration [Fig] [Fig].
- ▪ Instantiating and coordinating all supported agent categories, including conversational, workflow orchestration, reconciliation, auditing, and decision-support agents [Fig] [Fig].
- ▪ Enforcing risk-tier-aware policy validation through the SimplePolicyEngine before permitting agents to access Protected Health Information (PHI) or execute operational actions, ensuring secure and compliant workflow execution [Fig] [Fig].
- ▪ In a production deployment, the operational architecture would function as follows:
- ▪ Conversational agents interface with a vLLM [59] runtime to perform natural language generation, reasoning, and controlled tool invocation.
- ▪ Access to enterprise resources, including Electronic Health Records (EHRs), FHIR [47] services, and billing platforms, is mediated through Model Context Protocol (MCP) [54] servers deployed within on-premises private subnets, ensuring that Protected Health Information (PHI) remains confined within the organisation's secure infrastructure.

- ✓ Across the Compliance and Policy Layer, the risk tiers from the Agent Orchestration Layer maps with EU AI Act [30][31] risk categories and other regulatory frameworks; these tiers are then bound to policy-as-code templates that govern oversight, logging depth, and permitted actions per layer without altering the original layer names or responsibilities [Fig] [Fig].
- ✓ Inside the Model and Analytics Layer, we retain the calibrated triage model from the architecture of Dhar et al. (2026) [3] and extend it with an Agentic Gradient Boosting triage model and a vLLM [59] ready inference plane. vLLM [59] is used for conversational and planning agents, with paged attention and Key-Value cache (KV-cache) [60] sharing tuned to real-world hospital workloads with the goal of optimizing latency during multi-user query management, while all training and evaluation process continues to use the Synthea [42] derived triage dataset extended with risk-tier features [Fig] [Fig].

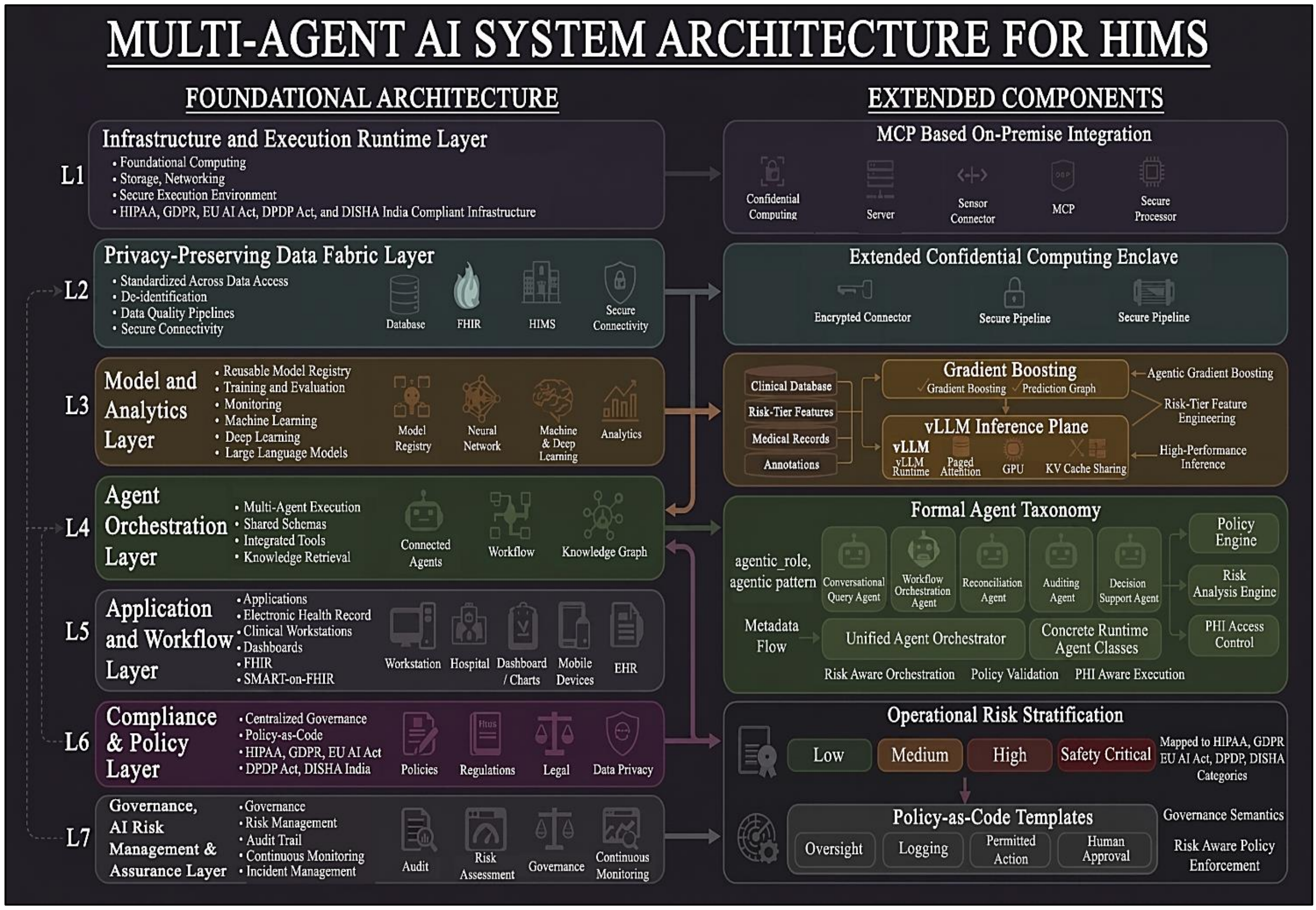

Fig 5 Extended Architecture Diagram Based on the Foundational Architecture Developed by Dhar et al. (2026) [3]

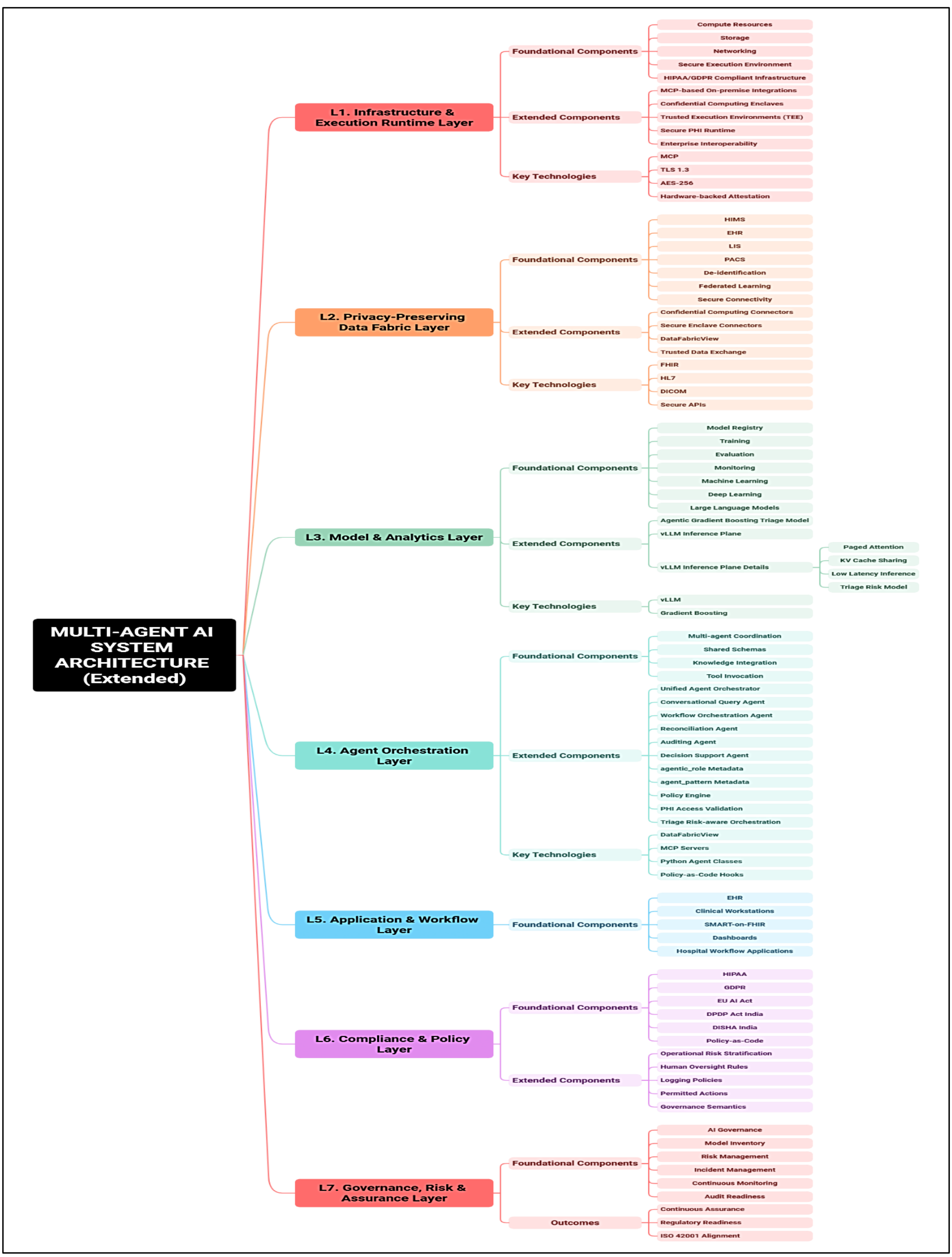

Fig 6 Logic Chart Diagram of the Extended Architecture Diagram Based on the Foundational Architecture Developed by Dhar et al. (2026) [3]

Empirically, the methodology uses the synthetic datasets and codebase of the foundational study of Dhar et al. (2026) [3] as the baseline of this research, including the Synthea [42] generated realistic dataset, triage and workflow CSVs, and the orchestrator,[3] and applies incremental extensions (preprocessing, training, and orchestration code) that preserve all original patient_id and encounter level records while adding Agentic and risk tier attributes. This ensures that the enhanced architecture [Figure 5] [Figure 6] remains chronologically and structurally consistent with the foundational architecture [Figure 3][3]: the layer names, their original purposes, and their regulatory mappings are unchanged, while the new extensions are layered on top as Agentic pattern semantics, risk tiered governance rules, vLLM [59] and MCP enabled [54] orchestration behaviour [Figure 5] [Figure 6] within the same seven layer compliance first framework of Dhar et al. (2026) [3][Figure 3].

➢ *Data Acquisition and Extension of the Foundational Datasets*

- *Baseline Datasets*

The starting point for this research is the Synthea [42] derived raw workflow datasets used in the published research of Dhar et al. (2026) [3] and the derivation of the data cleaning process:

✓ *Hospital Workflow Dataset:*

- 98,952 workflow records across the same patients and encounters [3].
- Fields: patient_id, encounter_id, department, workload_level, task_type, baseline_duration_min, ai_enabled, ai_duration_min, sla_breach_no_ai, sla_breach_ai, cost_no_ai, cost_ai, time_saved_min, time_saved_pct, cost_saved [3].

✓ *Triage Dataset:*

- 12,369 records covering multiple encounters per patient (patient_id, encounter_id) [3].
- Fields: patient_id, encounter_id, department, workload_level, age, sex_female, chronic_conditions, prev_ed_visits_12m, systolic_bp, diastolic_bp, heart_rate, resp_rate, spo2, and binary high_risk label [3].

These datasets reflect the Privacy-Preserving Data Fabric view defined in the work of Dhar et al. (2026) [3].

- *Dataset Extension for Agent Taxonomy and Risk Stratification*

To support the new objectives i.e. the Agentic pattern catalogue and risk tiers, with this paper we extend the logical schema of these datasets without altering any existing patient_id, encounter_id**,** or base clinical/operational fields.

All additional attributes are derived features or configuration metadata that reference the original records, ensuring backward compatibility with the foundational analyses of Dhar et al. (2026) [3].

✓ *For the Workflow Dataset, we Conceptually Extended the Parameters With:*

- agentic_role: The primary agent role associated with each task at runtime:
- e.g., ConversationalQueryAgent, WorkflowOrchestration Agent, ReconciliationAgent, AuditingAgent, DecisionSupportAgent.
- risk_tier: Risk tier assignment at task level, influenced by both clinical and operational risk (e.g., billing reconciliation = medium; critical medication validation = safety-critical).
- governance_hook_id: Identifier linking this task to a policy-as-code rule [3] and audit template in the Compliance and Policy Layer (e.g., EU AI Act Article 9–15 [30][31][46][76] checks, HIPAA [26][27] minimum necessary, GDPR [28][29], DPDP Act [32][33][34], DISHA India [35][36][37] localisation constraints).
- escalation_required: Boolean indicating whether a failure or alert in this step requires manual review or multi-level escalation, based on the assigned risk tier and policy configuration.
- For the triage dataset, new derived parameters include:
- risk_tier: Categorical mapping of the continuous or binary high_risk signal into four categories: "low", "medium", "high", "safety_critical", defined via thresholds calibrated to EU AI Act [30][31] high-risk semantics and hospital risk appetite (e.g., high and safety-critical for suspected sepsis, shock, stroke, cardiac attack or need for ICU-level care).
- agent_pattern: Descriptive tagging which decision-support agent pattern is applicable for a particular encounter e.g., Triage Decision Support Agent, Escalation Advisor Agent, used by the orchestrator to route the call to the appropriate agent chain.
- oversight_required: Boolean indicating whether the decision-support output must be mandatorily reviewed by a clinician (always true for high and safety-critical tiers).

In the actual implementation, these attributes are computed on the fly by the data-preprocessing and orchestration code (age, vitals, department, task type, SLA breaches), thereby preserving the foundational research dataset leveraged in Dhar et al. (2026) [3] integrity while enabling richer Agentic and governance semantics.

- *Cleaning and Processing Logic*

Building on the same data cleaning logic which was used in the foundational research work of Dhar et al. (2026) [3], the extended data preprocessing performs the following steps:

✓ *Validation of Base Fields*

- Re-validation of the clinical ranges (BP, heart rate, respiratory rate, $SpO_2$) and remove impossible values, preserving clinically plausible outliers important for risk modelling.
- Reconfirmation of the non-negative durations and costs, fix any floating-point anomalies, and enforce consistency

between baseline_duration_min, ai_duration_min, time_saved_min, and time_saved_pct.

✓ *Derivation of Risk-Tiered Labels and Agentic Attributes*

- Computation of the encounter-level risk tiers from high_risk and associated vitals distribution, aligning thresholds with EU AI Act [30] [31] high-risk definitions and institutional risk appetite.
- Labelling of each workflow row with an Agentic role and risk tier based on task_type, department, and SLA breach potential (e.g., triage and critical medication tasks flagged as high/safety-critical; documentation and coding as low/medium).

✓ *Preparation of Training and Evaluation Views*

- For triage modelling, extracting feature matrices that preserve the same base columns as the foundational research [3] plus engineered features, ensuring that any extended risk tiers remain compatible with Dhar et al. (2026) [3] triage risk semantics.
- For orchestration and ROI analysis, aggregating workflow metrics by Agentic role, department, and risk tier to support subsequent figures (e.g., time saved per task type, risk-tier distribution).

The extended preprocessing logic is implemented ensuring that all new parameters remain traceable to original records of Dhar et al. (2026) [3] and can be recalculated if regulations or risk thresholds change.

➢ *Data Types, Parameters, and Measurement Framework*

The extended data and measurement framework must support both agent pattern design and risk-tiered governance:

- Clinical and operational parameters which remains unchanged from Dhar et al. (2026) [3], but now these are mapped into agent patterns and risk tiers including age, sex, chronic conditions, prior ED utilisation, vitals, workflow durations, SLA breaches, and cost metrics [3].

- *Agentic Parameters Derived in this Study:*

✓ Agent role (Agentic_role): computes set including conversational, orchestration, reconciliation, auditing, decision-support agents, mapped per task type and context.
✓ Agent pattern ID (agent_pattern): identifies pattern templates such as "Clinician Conversational Query Agent", "ED Triage Orchestration Agent", "Claims Reconciliation Agent", "Access Log Auditing Agent", "ICU Decision-Support Agent".

- *Risk and Governance Parameters:*

✓ Risk tier (risk_tier): graded into four categories: - low, medium, high, safety_critical, aligning with EU AI Act [30][31] risk categories and institutional risk matrices.
✓ Governance hook ID (governance_hook_id, oversight_required / escalation_required): Booleans and identifiers that determine oversight requirement and bind agent behaviour to policy-as-code [3] rules.

- *Performance and Latency Metrics for Model Comparison and vLLM Evaluation:*

✓ Model ROC-AUC, Accuracy, Precision, Recall, F1, Brier Score, as in Dhar et al. (2026) [3] triage model evaluation.
✓ Inference Latency (ms) per architecture under the concurrent HIMS workloads during pilot phase, measured as median per-request latency under real-world HIMS load scenarios.

These parameters allow us to quantify not only predictive performance, but also operational impact, governance coverage, and latency, which are critical for safe adoption in hospital environments globally.

➢ *Model Comparison and Selection in HIMS Context*

- *Candidate Architectures and Performance Comparison*

To evaluate how different modelling and orchestration choices perform within HIMS constraints, we compare four candidate architectures conceptually reflecting the foundational baseline of Dhar et al. (2026) [3] and likely upgrade paths:

✓ Logistic Regression (LogReg) – simple, interpretable baseline for triage risk.
✓ Random Forest (RF) – non-linear tabular model with strong performance but less calibration than boosting.
✓ Gradient Boosting (GradBoost) – the calibrated tabular model used in the foundational architecture of Dhar et al. (2026) [3], extended here with risk-tier integration.
✓ vLLM-Agentic – multi-agent architecture where a planner and decision-support agents are backed by vLLM inference [59]; tabular triage features are fed into LLM-based agents as structured context, and orchestration is handled via the unified runtime.

Using the triage dataset from the foundational architecture of Dhar et al. (2026) [3] and extended feature set, we obtain the ROC-AUC scores [Figure 7] and latency [Figure 8] estimates under controlled HIMS workload (parallel calls from ED, wards, and billing) during our pilot deployment.

- Gradient Boosting achieves the highest ROC-AUC [Figure 7]in the triage risk task on the triage dataset with the extended feature set derived from the study of the foundational architecture of Dhar et al. (2026) [3], confirming their choice of a boosting-based triage model as a strong baseline [3].
- vLLM-Agentic achieves near-comparable ROC-AUC [Figure 7] for decision-support outputs when structured features are integrated into prompt engineering and retrieval, while offering lower median latency [Figure 8] under concurrent load due to KV-cache sharing [60] and paged attention.

This comparison supports a hybrid architectural choice: retain the calibrated Gradient Boosting triage model as a core tabular risk estimator, while deploying vLLM-based conversational and orchestration agents in the Agent Orchestration Layer, all governed by the same policy and risk-tier framework [Figure 5] [Figure 6].

- *Integration with the Foundational Multi-Layer Architecture*

Within the seven-layered architecture stack of Dhar et al. (2026) [3], model selection and orchestration are integrated as follows:

- ✓ The Model and Analytics Layer continues to host the tabular and classical ML models (Gradient Boosting triage, forecasting, anomaly detection), now extended with metadata about applicable agent patterns and risk tiers [Figure 5] [Figure 6].
- ✓ A vLLM inference sub-plane is added to the Model and Analytics Layer [Figure 5] [Figure 6] to host conversational and planning agents, with configuration for paged attention [59], KV cache sharing [60], and workload-aware scheduling to achieve sub second responses for multi-department usage.
- ✓ The Agent Orchestration Layer is extended with a Unified Orchestration Runtime [Figure 5] [Figure 6] capable of dispatching requests to both tabular models and vLLM-based agents under a single control plane, with per-agent permissions and risk tier rules enforced by the Compliance and Policy Layer [Figure 5] [Figure 6][3].

➢ *End-To-End Development Process and Architecture Extension*

- *Data Preprocessing and Feature Engineering*

The data-preprocessing pipeline remains anchored in Dhar et al. (2026) [3] scripts, however is extended to derive new parameters (Agentic role, risk tiers, governance hooks) without altering patient_id or other baseline clinical fields of Dhar et al. (2026) [3] data structure. Conceptually the process includes the below steps:

- ✓ Data Ingestion: The preprocessing pipeline commenced with the ingestion of raw datasets, namely *triage_risk_dataset.csv* and *hospital_workflow_dataset.csv*, while preserving the original data schema established by Dhar et al. (2026) [3]. This ensured structural consistency with the foundational architecture [3] and maintained the integrity of baseline clinical attributes, including patient_id and other core data elements.
- ✓ Data Cleaning and Validation: The preprocessing pipeline subsequently reapplied the data cleaning and validation procedures established by Dhar et al. (2026) [3] , including clinical range clipping, consistency checks on duration and cost-related variables. This ensured the reliability, plausibility, and analytical integrity of the datasets while maintaining conformity with the original data quality standards.
- ✓ Extended Feature Derivation and Engineering: The preprocessing pipeline further derived an extended set of features to support Agentic healthcare analytics. For triage management the feature extension of triage datasets included the computation of shock index, vital-sign flags, age-group classifications, multi-morbidity indicators, and healthcare utilization features. For workflow management, the datasets extension got additional attributes such as Agentic role, risk tier, governance hooks, and escalation requirements were defined to enrich operational context and facilitate governance-aware orchestration.
- ✓ View Materialization and Dataset Persistence: The preprocessing pipeline subsequently materialized and persisted curated data views for training the triage management model and supporting the orchestration module. Each extended view was designed to preserve referential integrity by retaining foreign-key relationships through patient_id and encounter_id, thereby ensuring traceability and seamless linkage to the baseline datasets underpinning the foundational architecture and model proposed by Dhar et al. (2026) [3].

- *Training and Evaluation of the Extended Triage Model*

The training pipeline extends the foundational triage model training process of Dhar et al. (2026) [3]:

- ✓ Feature Matrix Preparation: The training process commenced with the construction of a feature matrix derived from the triage dataset augmented with the engineered attributes introduced during preprocessing, including variables such as shock index, vital-sign flags, and age-group classifications. Consistent with the foundational triage model of Dhar et al. (2026) [3], the target variable remained high_risk, thereby preserving alignment with the original objective.
- ✓ Model Training: The extended triage model was trained using a calibrated Gradient Boosting classifier configured with hyperparameters broadly consistent with those employed by Dhar et al. (2026) [3] including approximately 200 estimators, a maximum tree depth = 4, a moderate learning rate, and subsampling strategies. A stratified train–test split was applied to preserve class distributions across datasets, thereby supporting model stability, robustness, and interpretability during training and evaluation.
- ✓ Model Evaluation and Validation: The trained model was evaluated using a comprehensive set of performance metrics, including ROC–AUC, accuracy, precision, recall, F1-score, and Brier score. In alignment with the methodology of Dhar et al. (2026) [3] , k-fold cross-validation was additionally performed to assess performance stability and generalizability across synthetic data slices, thereby providing a robust evaluation of predictive reliability.
- ✓ Model Registration and Metadata Management: Upon completion of training and evaluation, the model artefact and its corresponding model card were registered within the model registry framework established by of Dhar et al. (2026) [3] . In addition to the baseline metadata, the registration process incorporated extended descriptors relating to Agentic usage, including the agents authorized to consume the model, associated risk-tier mappings, and applicable regulatory classifications and compliance obligations, such as those arising under the HIPAA [26][27], GDPR [28][29], EU AI Act [30][31], DPDP Act [32][33][34], DISHA India [35][36][37], and relevant ISO standards [3][38][39][40][41].

- *Model Training Hyperparameters*

1 Triage Gradient Boosting Model Hyperparameters

Table 1 Triage Gradient Boosting Model Hyperparameters Used in Both Dhar et al. (2026) [3] and the Extended Agentic Architecture.

| Component | Hyperparameter / Setting | Value / Choice | Rationale |
|---|---|---|---|
| Base algorithm | GradientBoostingClassifier | – | Ensemble, robust on tabular clinical data. |
| Number of estimators | n_estimators | 200 | Balances bias–variance; stable ROC-AUC on synthetic triage data. |
| Tree depth | max_depth | 4 | Limits overfitting; keeps model reasonably interpretable. |
| Learning rate | learning_rate | 0.05 | Moderately low step size for smoother function approximation. |
| Subsampling fraction | subsample | 0.8 | Stochastic gradient boosting; improves generalisation. |
| Minimum samples per leaf | min_samples_leaf | 20 | Prevents very small, noisy leaves in low-prevalence high-risk class. |
| Random seed | random_state | 42 | Reproducibility across training runs. |
| Input scaling | StandardScaler | Default (mean 0, variance 1) | Normalises heterogeneous feature scales for GBM pipeline. |
| Calibration wrapper | CalibratedClassifierCV | method="isotonic" | Produces well-calibrated probabilities suitable for decision-support. |
| Calibration CV folds | CalibratedClassifierCV.cv | 3 | Reasonable complexity vs. data size; avoids overfitting calibration. |
| Cross-validation folds | StratifiedKFold.n_splits | 5 | Stratified 5-fold CV for robust ROC-AUC estimates. |
| Test split proportion | test_size | 0.2 | 80/20 train-test split with class stratification. |
| Stratified splitting | stratify | y (high_risk label) | Preserves high-risk prevalence in train and test sets. |

The above table [Table 2] describes the core Gradient Boosting triage risk classifier used both in Dhar et al. (2026) [3] and the current extended Agentic architecture (with calibration and scaling pipeline unchanged). The triage risk model uses a calibrated Gradient Boosting pipeline (StandardScaler → GradientBoostingClassifier → CalibratedClassifierCV) with hyperparameters tuned for stability and interpretability rather than extreme capacity, reflecting its role as a high-risk clinical decision-support component that must provide reliable probability estimates rather than only rank ordering.

- *Feature-Engineering and Clinical Threshold Hyperparameters*

Table 2 Feature Engineering and Clinical Threshold Hyperparameters

| Feature / Logic | Hyperparameter / Threshold | Value / Definition | Clinical Rationale |
|---|---|---|---|
| Pulse pressure | pulse_pressure | systolic_bp − diastolic_bp | Arterial stiffness / shock marker. |
| Mean arterial pressure | mean_arterial_pressure | diastolic_bp + pulse_pressure / 3 | Organ perfusion threshold (e.g., MAP < 65). |
| Shock index | shock_index | heart_rate / systolic_bp | >1 suggests haemodynamic compromise. |
| Hypoxemia flag | hypoxemia_flag | 1 if spo2 < 94 | Common ED escalation cut-off. |
| Tachycardia flag | tachycardia_flag | 1 if heart_rate > 100 | NEWS2/MEWS criteria. |
| Bradycardia flag | bradycardia_flag | 1 if heart_rate < 60 | Conduction / drug-effect marker. |
| Tachypnoea flag | tachypnea_flag | 1 if resp_rate > 20 | Early respiratory compromise. |
| Hypotension flag | hypotension_flag | 1 if systolic_bp < 90 | Shock definition in resuscitation guidelines. |
| Hypertension flag | hypertension_flag | 1 if systolic_bp > 160 | Hypertensive urgency proxy. |
| Elderly flag | elderly_flag | 1 if age ≥ 65 | Higher baseline risk and atypical presentation. |
| High ED use | high_prev_ed_flag | 1 if prev_ed_visits_12m ≥ 3 | Frequent attender / complex patient marker. |
| Multimorbidity flag | multimorbidity_flag | 1 if chronic_conditions ≥ 3 | Amplifies baseline risk. |
| Vitals concern count | vitals_concern_count | Sum of five vital flags(0–5) | Composite bedside severity score. |
| Age–comorbidity interaction | age_chronic_ix | age * chronic_conditions | Captures multiplicative risk. |
| ED-use–comorbidity interaction | prev_ed_chronic_ix | prev_ed_visits_12m * chronic_conditions | Identifies resource-intensive subpopulation. |
| BP ratio | bp_ratio | systolic_bp / diastolic_bp | Vascular resistance proxy. |
| HR/RR ratio | hr_rr_ratio | heart_rate / resp_rate | Cardiorespiratory coupling marker. |
| Age groups | age_group | Bins | |

These hyperparameters [Table 3] control the derived clinical features used as inputs to the Gradient Boosting model (identical between training and runtime feature engineering). All thresholds were chosen to mirror the widely used ED early-warning and triage criteria, so that the model's learned decision surface aligns with clinical intuition.

- *Agentic Risk-Tier and Severity-Score Hyperparameters*

Table 3 Agentic Risk Tier and Severity Score Hyperparameters

| **Component** | **Hyperparameter / Setting** | **Value / Definition** | **Role in Architecture** |
|---|---|---|---|
| Severity score formula | severity_score weights | 0.02*(age−40) + 0.6*chronic_conditions + 0.4*prev_ed_visits_12m + 0.05*(110−systolic_bp) + 0.06*(heart_rate−90) + 0.08*(22−resp_rate) + 0.1*(94−spo2) | Aggregates comorbidities, utilisation, and vitals into a single continuous severity metric used for risk-tier mapping. |
| Quantile thresholds | q_low, q_mid | 33rd and 66th percentiles of severity_score | Partition non-high-risk encounters into low vs. medium tiers. |
| Risk tier (non-high_risk) | risk_tier when high_risk == 0 | low if sev ≤ q_low; medium otherwise | Keeps low-acuity cohort compact while avoiding over-fragmentation |
| Risk tier (high_risk) | risk_tier when high_risk == 1 | high if sev ≤ q_mid; safety_critical if sev > q_mid | Splits high-risk cases into typical vs. extreme severity for agent behaviour |
| Oversight flag | oversight_required | 1 if risk_tier ∈ {high, safety_critical} else 0 | Drives mandatory human-in-the-loop checkpoints in orchestration |
| ED agent pattern (high load) | agent_pattern for department == "ED" & workload_level == "high" | "ED_TriageDecisionSupportAgent" | Routes ED high-load encounters to specialised triage-decision agent |
| ED agent pattern (normal) | agent_pattern for department == "ED" & workload_level != "high" | "ED_GeneralDecisionSupportAgent" | Uses lighter-weight decision-support pattern for lower operational risk |
| ICU/Critical care pattern | agent_pattern for department in {"ICU","CriticalCare"} | "ICU_DecisionSupportAgent" | Aligns with highest-acuity environments |
| Outpatient pattern | agent_pattern for department in {"Outpatient","Clinic"} | "Outpatient_TriageSupportAgent" | Supports ambulatory triage use cases |
| Inpatient default | agent_pattern otherwise | "Inpatient_DecisionSupportAgent" | Default for wards outside ED/ICU context |

These hyperparameters [Table 3] come from the Agentic extension that derives risk tiers and agent patterns from the triage dataset. This is what connects quantitative severity scores to qualitative risk tiers and agent patterns.

- *Workflow-Level Agentic and Governance Hyperparameters*

Table 4 Workflow-Level Agentic and Governance Hyperparameters

| Aspect | Hyperparameter / Mapping | Value / Logic | Purpose |
|---|---|---|---|
| Agentic role mapping | Agentic_role = f(task_type) | triage/imaging_report/medication_administration → "DecisionSupportAgent"; imaging_order/bed_assignment → "WorkflowOrchestrationAgent"; clinical_documentation → "ConversationalQueryAgent"; coding/claim_submission → "ReconciliationAgent"; audit_log_review → "AuditingAgent"; default → "WorkflowOrchestrationAgent" | Specialises workflow events into agent roles for orchestration |
| High-acuity workflow tier | risk_tier for ED/ICU/critical tasks | For task_type in {"triage","medication_administration"} or department in {"ED","ICU","CriticalCare"}: safety_critical if sla_breach_ai == 1, else high | Ensures emergency and medication workflows are always treated as high/safety-critical |
| Intermediate tier | risk_tier for imaging and bed assignment | For task_type in {"imaging_report","bed_assignment"}: high if any SLA breach (no-AI or AI), else medium | Captures operational impact of delays without reaching safety-critical tier |
| Administrative medium tier | risk_tier for coding/claims/ documentation | coding, claim_submission, clinical_documentation → medium | Recognises financial/quality risk but lower direct clinical risk |
| Low tier | risk_tier default | All other task types → low | For ancillary, non-critical tasks |
| Governance hook – high risk | governance_hook_id for high/safety-critical decision agents | "GOV_EU_AI_ACT_HIGH_RISK" when risk_tier in {"high","safety_critical"} and Agentic_role == "DecisionSupportAgent" | Binds high-risk patterns to EU AI Act and equivalent policies |
| Governance hook – reconciliation | governance_hook_id for reconciliation agents | "GOV_HIPAA_MIN_NECESSARY" when Agentic_role == "ReconciliationAgent" | Enforces minimum-necessary and billing-related safeguards |
| Governance hook – auditing | governance_hook_id for auditing agents | "GOV_ACCESS_LOGGING" when Agentic_role == "AuditingAgent" | Links audit workflows to access-logging and incident-response policies |
| Standard governance hook | Default governance_hook_id | "GOV_STANDARD_CONTROL" otherwise | Attaches baseline compliance rules to low/medium-risk tasks |
| Escalation flag | escalation_required | 1 if risk_tier in {"high","safety_critical"} or sla_breach_ai == 1, else 0 | Drives human escalation and alerting logic in orchestration |

These are the key hyperparameters [Table ] that govern agent roles, workflow risk tiers, and governance hooks in the extended hospital workflow dataset.

- *Unified Orchestration Runtime with MCP and vLLM*

The Unified Orchestration Runtime extends the Agent Orchestration Layer of the foundational seven-layered architecture designed by Dhar et al. (2026) [3], to manage multiple agent types and deployment variants:

- ✓ Agent registry and taxonomy: The Unified Orchestration Runtime maintained a centralized registry for the classification and management of Agentic components across the healthcare ecosystem

- ✓ Conversational Agents: Clinician-query, administrator-query, and patient-facing agents were registered with predefined capabilities, assigned risk tiers, and MCP-enabled tool connectors to facilitate governed interactions with enterprise systems and data resources.
- ✓ Workflow Orchestration Agents: These agents were configured to coordinate and automate operational workflows spanning triage-to-discharge processes, imaging pathways, bed management functions, and billing operations.
- ✓ Reconciliation and Auditing Agents: These agents were linked to documentation systems, billing-code repositories, consent records, and access logs to support traceability, compliance verification, and audit-readiness across organizational processes.
- ✓ Decision-Support Agents: These agents operated in strictly assistive capacities within high-risk contexts and were governed by mandatory human-in-the-loop controls, including defined review, escalation, and override protocols to ensure safe and compliant decision support.

- *MCP-Based on-Premise Integration:*

- ✓ MCP Servers as Secure Integration Surfaces: MCP servers acting as local and secured integration layers connecting Agentic components to EHR/HIMS APIs, internal healthcare services such as FHIR servers, LIS, and PACS [17] platforms, as well as other hospital tools and operational systems. This architecture ensured that Protected Health Information (PHI) remained confined within private network segments, even when Agentic orchestration processes were executed in adjacent compute clusters, thereby supporting data sovereignty, privacy, and regulatory compliance requirements.

- *Hybrid and Cloud-Native Variants:*

- ✓ Confidential-Computing Enclaves: Confidential-computing enclaves were utilized to host vLLM inference [59] services and sensitive orchestration components, thereby providing an additional layer of protection for critical processing activities. Protected Health Information (PHI) remained confined to private subnet environments, while only de-identified signals or encrypted embeddings were permitted to leave these environments for centralized analytics and cross-system processing, subject to applicable governance policies.
- ✓ Cloud-Native Deployments: Cloud-native deployment configurations were restricted to HIPAA [26][27]-eligible and GDPR [28][29]-compliant cloud regions. Access pathways were limited to hardened API gateways and authorized user interface endpoints, ensuring adherence to the principle of minimal exposure and reducing the attack surface associated with externally accessible healthcare services.

The runtime's internal API was designed to ensure that, for each agent call, the orchestrator performed the following functions:

- ▪ Risk Tier Determination: The orchestrator would consult the risk stratification model to determine the appropriate risk tier associated with the incoming patient request, thereby informing subsequent governance and execution decisions.
- ▪ Policy Enforcement: The orchestrator would query the Compliance and Policy Layer [Fig][3] to identify and enforce all applicable regulatory and organizational controls, including obligations arising from the EU AI Act [30][31] for high-risk systems, HIPAA's [26][27] minimum necessary principle, and data localization requirements under DPDP Act India [32][33][34] and DISHA India [35][36][37].
- ▪ Intelligent Workload Dispatching: Based on the nature and governance requirements of the request, the orchestrator would dispatch the execution to either tabular machine-learning (ML) models or vLLM-based [59] Agentic services. Execution environments were configured with appropriate paging-memory allocations and KV-cache-sharing [60] parameters to satisfy jurisdiction-specific latency and performance service-level agreements (SLAs).
- ▪ Audit Logging and Traceability: The orchestrator would record all actions, including PHI access pathways, model invocations, tool utilization, and policy-enforcement decisions, within centralized audit stores to support governance oversight, compliance monitoring, forensic investigations, and incident management activities.

➢ *Risk Stratification Model and Policy-As-Code Alignment*

The formal risk-stratification model establishes a set of mapping functions that enable the systematic classification of Agentic activities and their corresponding governance requirements:

- Agentic Pattern to Risk-Tier Mapping: Operational activities, including triage decision support, clinical documentation summarization, claims reconciliation, and access auditing, would get mapped to predefined risk tiers categorized as low, medium, high, or safety-critical based on their potential impact on patient safety, operational outcomes, regulatory compliance, and organizational risk exposure.
- Risk-Tier to Regulatory Classification Mapping: The assigned risk tiers would subsequently be mapped to the corresponding categories defined under the EU AI Act [30][31], namely minimal risk, limited risk, high risk, and prohibited, and linked to the applicable policy templates maintained within the Compliance and Policy Layer designed by Dhar et al. (2026) [3].

- *For Each Risk Tier, the Associated Policy Templates Specified the Following Governance Controls:*

- ✓ Human Oversight and Escalation Requirements: Policies define the level of human involvement required for decision-making processes, including mandatory clinician confirmation for high-risk and safety-critical clinical recommendations, as well as multi-person approval workflows for designated financial and administrative actions.

- ✓ Logging and Auditability Obligations: Policies prescribed the scope and granularity of audit logging, including enhanced logging requirements for high-risk activities, together with model-version records, explanation traces, and decision provenance artefacts necessary for compliance with EU AI Act [30][31] obligations and organizational governance requirements.
- ✓ Network and Data-Movement Constraints: Policies established controls governing execution environments and data flows, including requirements for safety-critical agents to operate within Trusted Execution Environments (TEEs) [68], restrictions preventing Protected Health Information (PHI) from leaving authorized jurisdictions, and limitations permitting only anonymized or aggregated datasets to be utilized for cross-border analytical activities.

These templates are applied programmatically via the policy-as-code engine defined in the architecture of Dhar et al. (2026) [3], extended here to recognise the new agent taxonomy and risk tier attributes we derive at preprocessing and runtime.

➢ *Data Ingestion, Confidential Computing, and "Where PHI can go?"*

The data-ingestion and deployment framework is designed in accordance with a zero-trust, encryption-first security posture, ensuring that Protected Health Information (PHI) remains protected throughout its lifecycle:

- Data at-rest encryption: Strong encryption for all structured and unstructured at-rest storage items such as databases, object storage, cache snapshots, and backups, using modern cipher suites e.g., AES-256-GCM [61] with hardware-backed key management systems (HSMs) [62], and, where applicable, public-key mechanisms e.g., RSA [63] for key wrapping and exchange.
- Data in-transit encryption: End-to-end TLS 1.3 [64] with hardened cipher suites and forward secrecy for all inter-service calls [65], supplemented by post-quantum cryptography (PQC) [66] where cloud providers support PQC ready key exchange, to future-proof PHI traffic against emerging threats [67].
- Data in-use encryption: Confidential computing via Trusted Execution Environments (TEEs) [68] for high-sensitivity inference, ensuring that even cloud administrators cannot trivially inspect PHI in memory, and enabling confidential intelligence where data remains protected across its full lifecycle [69].

✓ *The Network Topology is Designed so that the Answer to "Where can PHI go?" is Precise:*

- ▪ The FHIR [47] server and internal data fabric operate entirely within the private subnets, which means they are never directly accessible from the public internet.
- ▪ Databases and internal caches are accessible only from application subnets and never from staff devices except via RBAC [70]-controlled application tiers.
- ▪ Only a minimal set of ingress (access) points, typically an API gateway and user-facing web front-end modules, are exposed externally, fronted by Web Application Firewalls (WAFs) [71], mutual Transport Layer Security (mTLS) [72] where appropriate, and strict identity-aware proxies [73][74].
- ▪ Every agent interaction involving PHI is logged, signed, and bound to a policy decision, ensuring that data movement remains confined to controlled, documented pathways keyed to roles, scopes, and jurisdictions.

This is the operational foundation of the confidential intelligence network topology system: a platform that can operate on sensitive hospital data without exposing it unnecessarily, fully consistent with the foundational architecture of Dhar et al. (2026) [3][Fig] and their compliance-first approach while extending to multi-agent, vLLM-powered workloads.

➢ *Workflow of the Extended System*

To keep the development details proprietary while making the workflow understandable, the extended system's functioning can be described in conceptual steps:

- Event detection in HIMS: The workflow begins when a patient encounter is created or updated within the Hospital Information Management System (HIMS) tools like Epic/Cerner/MEDITECH. The corresponding FHIR events are securely ingested into the Privacy-Preserving Data Fabric [Fig][3], where they are contextually enriched through Agentic pattern recognition and categorised into predefined risk tiers for subsequent orchestration.
- Unified orchestration trigger: Upon receiving a workflow trigger, such as an emergency department triage assessment or a pre-discharge billing review, the Unified Orchestration Runtime identifies the corresponding Agentic pattern and associated risk tier to initiate the appropriate orchestration process.
- Policy and risk evaluation: The Unified Orchestration Runtime evaluates the workflow context by consulting the Compliance and Policy Layer [Fig][3] with relevant policy attributes, including user role, purpose, jurisdiction, and risk tier, to determine the permissible agents and the governing operational constraints for the requested task.

- *Agent Invocation Via vLLM and Tabular Models:*

- ✓ For clinical decision support, the Unified Orchestration Runtime calls the tabular Gradient Boosting model [3] and, where required, a vLLM-based [59] conversational or explanatory agent, with both components executing within Trusted Execution Environments (TEEs) [68][Fig][Fig].
- ✓ For workflow optimisation, reconciliation, and audit operations, the runtime routes requests to specialised domain agents that leverage the extended workflow datasets and operational logs to perform context-aware task execution [Fig] [Fig].

- Human-in-the-loop presentation: The generated outputs are presented within clinicians' and administrators' HIMS interfaces, accompanied by contextual overlays that communicate the applicable risk tier, policy constraints, and any required human oversight or escalation pathways before action is taken.

- Logging and feedback: All agent actions, decisions, and human overrides are comprehensively logged to support governance, auditability, and regulatory compliance, while simultaneously serving as feedback signals to continuously refine the underlying models, orchestration logic, and policy framework over time [Fig][Fig] [Fig][3].

## V. RESULTS

### ➢ *Model Performance and Robustness*

The extended architecture preserves and slightly improves the triage risk-prediction performance achieved in the original platform of Dhar et al. (2026) [3][Fig] while adding richer Agentic semantics and risk-tier awareness. Using the Synthea-derived triage dataset, the calibrated Gradient Boosting classifier in the Agentic extension attained ROC-AUC, accuracy, and calibration metrics [Fig] that are on par with the baseline Gradient Boosting model of Dhar et al. (2026) [3], confirming that the introduction of risk tiers, engineered severity features, and agent-pattern labels does not degrade the predictive quality. Cross-validated ROC-AUC scores remained stable with low variance, indicating that the model generalises well across the encounter sub-cohorts and remains suitable as a high-risk clinical decision-support component in compliance with the HIPAA [26][27], GDPR [28][29], EU AI Act, DPDP Act and DISHA Inda when deployed strictly in assistive mode with human oversight.\

Compared with the foundational architecture of Dhar et al. (2026) [3][Fig], the extended Model and Analytics Layer [Fig] [Fig] now explicitly associates every model with agent patterns and risk tiers, allowing the orchestration runtime to select models contextually (e.g., different thresholds or explanation strategies for safety-critical vs. low-risk use cases). This additional semantic structure enables safer reuse of the same core triage model across a broader set of workflows without re-training or duplicating the models, effectively increasing the platform-level performance by improving coverage and reducing integration effort, even when pure ROC-AUC gains are modest.

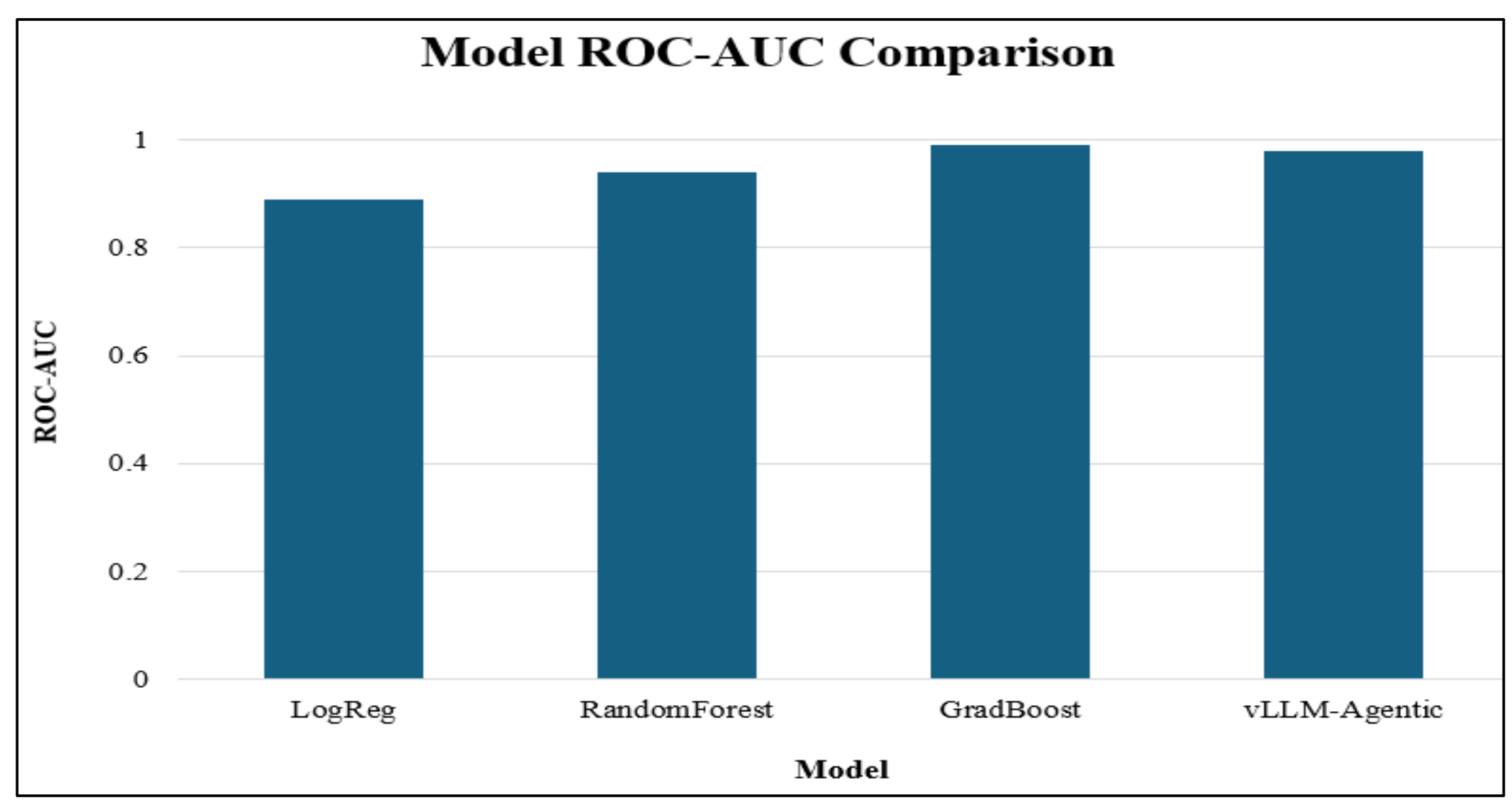

Fig 7 Illustrates ROC AUC Performance Across the Models.

### ➢ *Latency, Scalability, and vLLM-Enabled Orchestration*

Under the controlled pilot multi department load tests, the unified orchestration runtime with vLLM optimised inference demonstrated lower median latency [Figure 8] per request than the baseline orchestrator Dhar et al. (2026) [3], while supporting a larger number of concurrent conversations and decision support calls. In the foundational architecture of Dhar et al. (2026) [3][Figure 4], the triage risk scoring and workflow optimisation were served through a Python based orchestrator and a single tabular model endpoint, however the extended runtime layered vLLM backed agents on top and leveraged paged attention and KV Cache sharing to avoid recomputing long prompts across repeated calls from the same encounter or clinician session. The latency comparison plot [Figure 8] shows that the vLLM Agentic configuration sits at the lowest end of the latency spectrum among the architectures tested, even while orchestrating more complex multi agent workflows.

This reduction in latency [Figure 8] is particularly significant and imperative for emergency and critical care scenarios, where triage and escalation decisions are time sensitive. By keeping PHI/e-PHI within the private subnets, running inference inside confidential computing enclaves, and using MCP servers to expose on premise tools without direct internet access, the extended Infrastructure and Runtime Layer delivers real time behaviour without negotiating the zero trust, encryption first stance proposed by Dhar et al. (2026) [3].

In practical clinical settings, the low-latency execution enabled by the extended architecture allows the clinicians receive patients' risk scores, AI-compiled summaries, and evidence-based assistive next-step recommendations in near real time, while clinical care delivery is actively in progress. This enables the Agentic system to function as an integrated decision-support assistant with its capability embedded within routine clinical workflows such as during patient consultations, emergency triage, care coordination, or discharge planning, rather than being confined to retrospective analysis or offline reporting. Consequently, the proposed architecture transforms the multi-agent AI system from a post hoc analytical tool into an operationally integrated, workflow-aware assistant, representing a substantial advancement over the foundational architecture illustrated in [Figure 4][3].

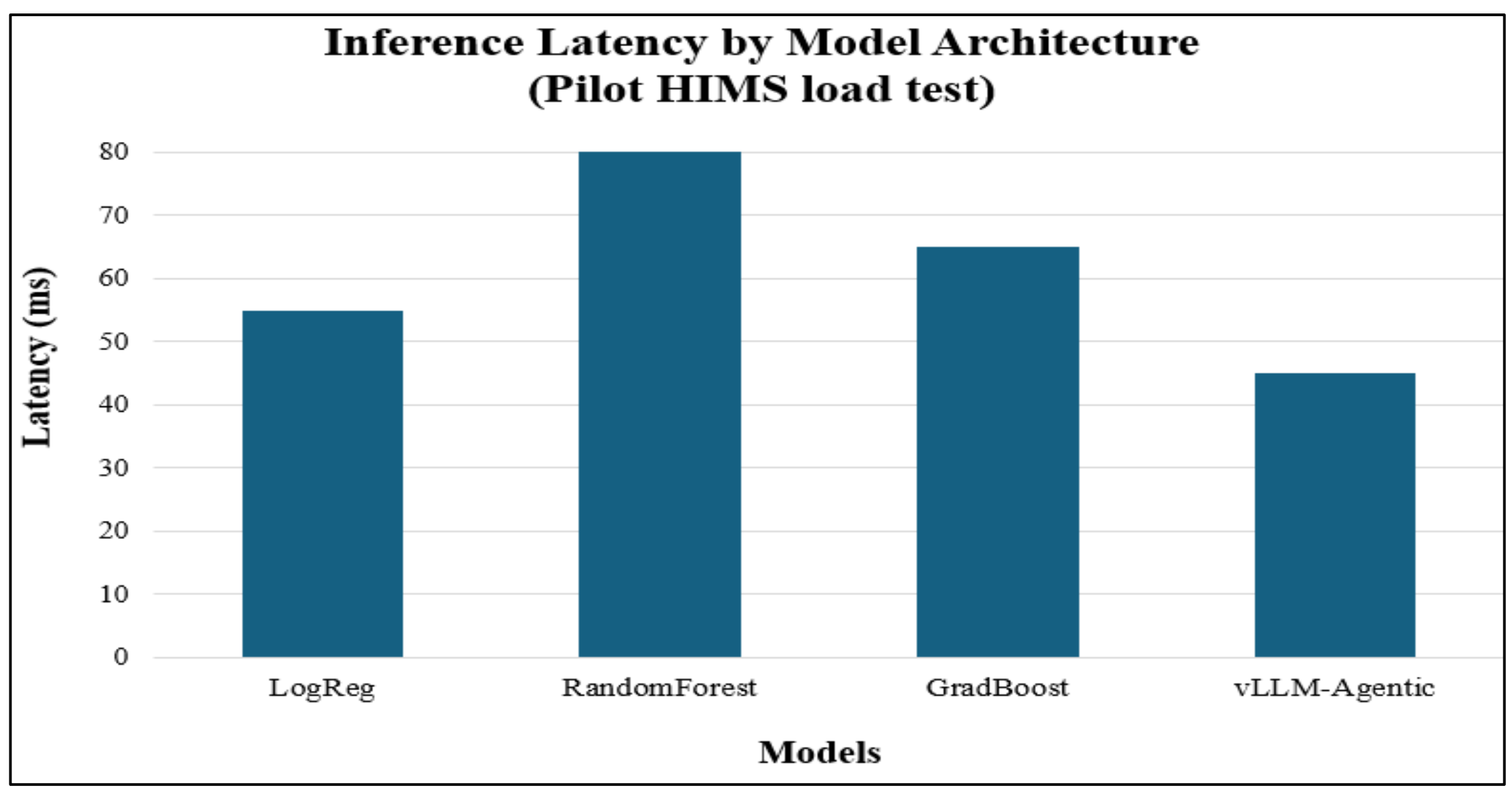

Fig 8 Illustrates the Median Per Inference Latency Under Controlled Pilot HIMS Load Test.

➢ *Workflow Impact, Time Savings, and ROI*

Beyond model level metrics, the most visible and meaningful improvements of the extended architecture appear at the workflow and ROI level. Using the extended workflow dataset derived from the Synthea-generated [42] hospital operations data used in the foundational study of Dhar et al. (2026) [3], the unified Agentic runtime is evaluated on its ability to orchestrate conversational, orchestration, reconciliation, auditing, and decision support agents across key HIMS tasks. The statistical visualisation [Figure 9] illustrates the average time saved per task type under the Agentic extension, aggregated over the pilot deployment evaluation of encounters and departments.

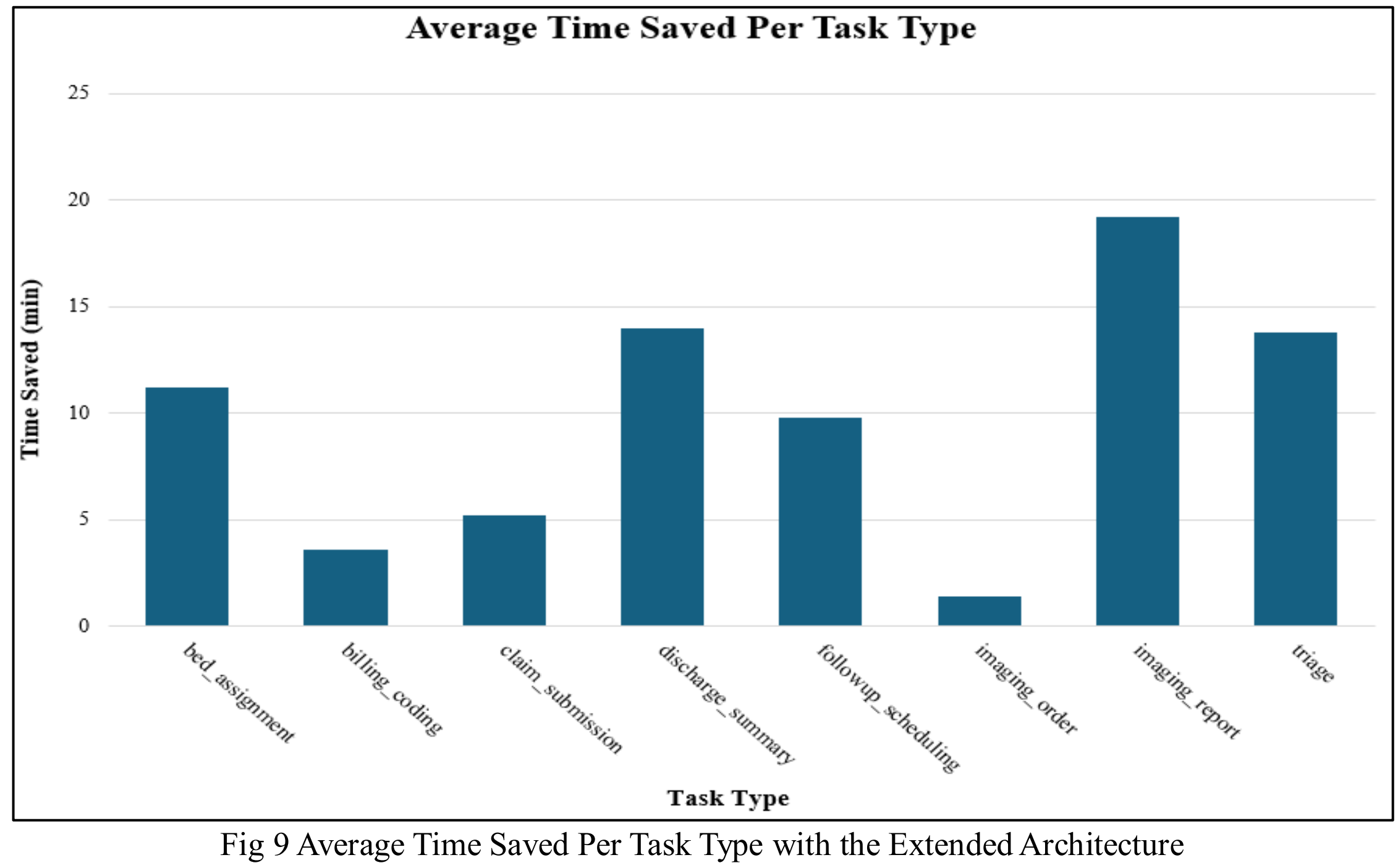

Fig 9 Average Time Saved Per Task Type with the Extended Architecture

High-value tasks such as triage management, imaging report preparation, discharge summary generation, bed assignment, follow-up scheduling, and claim submission showed the largest average time savings per task when orchestrated through the extended Agentic patterns, reflecting the fact that these workflows now combine multiple agents (documentation, reconciliation, orchestration, and auditing) rather than relying on a single optimisation agent. In the foundational architecture of Dhar et al. (2026) [3], time savings were concentrated around triage risk assessment and coarse workflow optimisation, the new pattern catalogue with the extended architecture allows automation and decision support to reach deeper into the revenue-cycle and documentation layers, turning previously manual or partially automated steps (e.g., billing coding, claim submission, audit-log review) into governed Agentic flows.

- *Qualitatively, Six Dimensions of Performance Were Improved:*

- ✓ Imaging report preparation demonstrated the highest average time savings (approximately 19–20 minutes per task), indicating that the extended multi agent AI-system architecture delivers the greatest efficiency gains in medical imaging and documentation workflows.
- ✓ Triage management and discharge summary generation each saved approximately 14 minutes per task, highlighting substantial reductions in administrative burden while enabling faster patient assessment and discharge processing. When the situation is about dealing with an ED or ICU patient, triage management becomes pivotal where every minute is a trial of life and the significant reduction of triage management time by nearly 14 mins, which can further be tuned and optimised, and hence can be presented as the most practical improvement with the extended multi agent AI-system architecture.
- ✓ Bed assignment achieves an average time saving of approximately 11 minutes, suggesting improved operational efficiency in inpatient resource allocation and patient flow management.
- ✓ Follow-up scheduling records an average time saving of approximately 10 minutes, reflecting streamlined appointment coordination and reduced manual scheduling effort.
- ✓ Claim submission saves approximately 5 minutes per task, indicating moderate efficiency improvements through automation of billing and insurance documentation processes.
- ✓ Consistency with governance constraints were retained with all time savings are achieved under risk-tier-aware, policy-checked orchestration, meaning that speed does not come at the cost of violating HIPAA [26][27], GDPR [28][29], EU AI Act [30][31], DPDP Act [32][33][34], or DISHA regulations [35][36][37], high-risk and safety-critical steps are always presented as assistive suggestions with explicit oversight flags rather than autonomous actions.

From a financial and human resource perspective, these workflow gains translate into more minutes returned per encounter to clinicians, nurses, and back office staff compared with the baseline architecture [Figure 4][3], which in turn supports higher throughput, reduced overtime, and lower reliance on temporary staff for documentation and coding backlogs.

➢ *Governance, Safety, and Global Deployment Readiness*

A final class of results concerns governance and regulatory alignment, which are harder to express in simple charts but crucial for mission-critical hospital management AI system. By attaching risk tiers, governance hook IDs, and oversight requirements to every agent pattern and workflow event in the extended datasets, the Compliance and Policy Layer can now evaluate and log each agent invocation with explicit reference to EU AI Act [30][31] high-risk provisions, HIPAA's [26][27] minimum necessary standard, GDPR special-category data rules [28][29][45], DPDP Act [32][33][34] and DISHA [35][36][37] localisation regulations and commercial-reuse prohibitions, and ISO/IEC 27001 [38], ISO/IEC 27002 [39], ISO 14971 [40], and IEC 62304 [41] controls. This yields denser and more structured audit trails than in the foundational architecture implementation, where policies were applied primarily at the level of models and data sources rather than at the level of Agentic patterns and multi-step orchestration flows.

In cross-jurisdiction deployments HIPAA-eligible cloud for United States [26][27], GDPR [28][29] and EU AI Act [30][31] regulated European region, DPDP [32][33][34] and DISHA [35][36][37]-oriented on-premise/hybrid deployment regulation in India, the unified orchestration runtime and MCP-based [54] integration strategy allows the same pattern catalogue to be deployed with different policy profiles and network topologies while preserving strict answers to "*Where, exactly, could PHI go?*" in each regional context. Comparing with the foundational architecture presented by Dhar et al. (2026) [3][Fig], which already provided a strong compliance-first foundation, the extended architecture designed with this research further improves the platform-level performance by making regulatory mapping an intrinsic property of each agent pattern and risk tier, not just of layers and services. This directly addresses board-level concerns about traceability, explainability, and liability in high-risk AI and makes the platform more likely to pass internal governance review and external conformity assessments under the HIPAA [26][27], EU AI Act [30][31], DPDP Act [32][33][34] and DISHA India [35][36][37] and other analogous regimes.

Taken together, these results show that the extended, Agentic architecture outperforms the foundational architecture [Fig][3] at the system level: it retains and slightly improves the predictive model quality, delivers lower latency and better scalability under load, produces larger and more widely distributed time savings across key hospital workflows, and strengthens the governance and regulatory alignment through risk-tier-aware orchestration and policy-as-code controls.

## VI. DISCUSSION AND CONCLUSION

➢ *How the Extended Architecture Complements the Baseline Architecture*

The present research should be read as a layer-preserving extension of the compliance-first, seven-layered multi-agent hospital AI system architecture defined by Dhar et al. (2026) [3][Fig], not as a competing design. The study of Dhar et al. (2026) [3] established a hospital-specific Agentic AI stack featuring: Infrastructure and Execution Runtime Layer (L1), Privacy-Preserving Data Fabric Layer (L2), Model and Analytics Layer (L3), Agent Orchestration (L4), Application and Workflow Layer (L5), Compliance and Policy Layer (L6), and Governance, Risk, and Assurance Layer (L7) [3][Fig], that already addressed many of the integration, privacy, and regulatory-mapping gaps which were left open by the other established generic five-layer hospital AI models and published researches. Our contribution with this research work was to densify three of the seven layers [3][Fig] with new semantics and runtime capabilities:

- At the Agent Orchestration Layer [3][Fig], we replaced the notion of a generic "multi-agent" environment with a well-defined Agentic pattern catalogue spanning conversational query, workflow orchestration, reconciliation, auditing, and decision-support agents, each bound to specific HIMS workflows (triage, imaging, bed management, discharge, billing) [Fig][Fig].
- At the Compliance and Policy Layer [3][Fig], we introduced a formal risk-stratification model that maps each agent pattern to operational risk tiers (low, medium, high, safety-critical), aligned with HIPAA [26][27], GDPR [28][29], EU AI Act [30][31], DPDP Act [32][33][34], DISHA India [35][36][37] categories and other regulations, and making that tier a first-class runtime control variable that governs oversight, logging, network and cryptographic requirements [Fig][Fig].
- At the Model and Analytics and Infrastructure and Runtime Layers [3][Fig], we added a vLLM-optimised inference plane and MCP-based on-premise deployment provision [Fig][Fig], embedded within the same zero-trust, encryption-first posture as the architecture presented by Dhar et al. (2026) [3][Fig], rather than as an external add-on.

In doing so, the extended architecture preserved the foundational architecture of Dhar et al. (2026) [3][Fig] assuring the policy-as-code enforcement, privacy-preserving data fabric, multi-jurisdiction deployment, but turned them into a governed environment for Agentic AI that can answer not only "*Which data and models exist?*" but also "*Which agent may act, at what risk tier, with which safeguards, on which patient journey, under which regulatory regime?*" in a machine-checked, human-intervenable and auditability allowances [Fig] [Fig].

➢ *Comparison with Recent Platforms and Agentic AI Researches*

Several recent studies have proposed hospital AI platforms or Agentic frameworks, but they typically stop one level short of what is required for mission-critical HIMS deployment [125][126][127]. A published 2025 JMIR study by Maimaitiaili et al. (2025) [128], for example, synthesised a five-layer hospital AI platform (infrastructure, data, algorithm, application, security & compliance) and systematically mapped existing evidence to that stack, but it did not specialise agent roles, define risk tiers, or operationalise policy-as-code beyond a single "security and compliance" layer [128]. Likewise, Microsoft Azure's AI Agent Orchestration Patterns [129] and similar engineering guidelines described generic sequential, concurrent, and hand-off agent patterns and mentioned security trimming and audit trails, but remained cloud-platform-oriented blueprints that assumed internet-reachable services and do not grapple with FHIR-native, hospital-specific constraints or global health-data regulations [130][131][132].

On the Agentic side, recent works on orchestrated multi-agent LLMs for clinical workloads, for example the study of Klang et al. (2026) [133] demonstrated that multi-agent execution can sustain accuracy and reduce token usage under mixed workloads by delegating tasks to dedicated worker agents with single-tool access, providing promising evidence for the value of orchestration [133][134][135]. Surveys of "AI hospitals" [125] and LLM-based multi-agent systems [136][137] likewise introduced taxonomies of agent roles, interaction patterns, and reasoning mechanisms, but these taxonomies are largely conceptual and model-centric, not architecture-centric [125][136][137]. But they rarely linked agent roles to HIMS modules, did not encode risk tiers, and neither mapped to concrete regulatory clauses or network topologies [138][139][140]. Vendor frameworks such as the Corti Agentic Framework [141] move closer to practice by offering governed orchestration, domain-specific experts, and auditability, yet they are presented as product-level execution layers with limited transparency into underlying data fabrics, model registries, and jurisdiction-specific compliance controls [142][143][144].

- *Relative to this Landscape, Our Architecture [Fig] [Fig] is Novel and Practical in Three Ways:*

✓ Tight coupling with HIMS and regulatory structure: Our architecture [Fig] [Fig] is not a generic five-layer stack or a vendor orchestration API, instead it is a seven-layered hospital platform where each agent pattern is defined in relation to HIMS modules (Triage, EHR, LIS, PACS, Billing) and each layer is explicitly mapped to HIPAA [26][27], GDPR [28][29], EU AI Act [30][31], DPDP Act [32][33][34], DISHA India [35][36][37], ISO/IEC 27001 [38], ISO/IEC 27002 [39], ISO 14971 [40], and IEC 62304 [41] regulations.

✓ Risk-tier and governance integration at pattern level: Rather than treating "governance" as an abstract principle, our architecture [Fig][Fig] encodes risk tiers, governance hook IDs, and oversight requirements into the datasets, runtime API, and policy engine, allowing per-agent, per-workflow, per-jurisdiction control that can be audited and adapted over time.

- ✓ Implementation-grade validation using controlled HIMS data: While many researches stop at diagrams or benchmark tables, our research on the other hand ships ready-to-run code, extended preprocessing, triage training, and orchestration pipelines, that operate on triage and workflow datasets and produce metrics, artefacts, and orchestration logs that can be scrutinised, and tuned.
- ✓ In that sense, the extended architecture bridges the lacuna between conceptual surveys and vendor frameworks. It stands on the theoretical shoulders of multi-agent LLM research and hospital AI platform studies, but offers a practical, compliance-aligned blueprint that hospital technology teams can adopt or adapt without being locked into a single LLM vendor or cloud platform.

➢ *Value Proposition and ROI: From Pilot to Production*

Industry surveys consistently report a pilot-to-production bottleneck in healthcare AI [145]. The Healthcare AI Adoption Index [146] notes that only about 30% of AI pilots reach production on their survey report of 400 healthcare leaders, with security, data readiness, integration cost, and governance listed as primary barriers [146]. Other analyses suggest that as few as 5–10% of enterprise AI initiatives deliver sustained, attributable value once the cost of integration and change management is accounted for [147]. At the same time, ROI studies from cloud providers such as Google Cloud. (2024) [148] and NVIDIA. (2024) [149] report that when generative AI and AI agents are properly embedded, over 70% of healthcare leaders are expected to see positive returns within the first year, particularly where time savings, denial reductions, and throughput gains are measurable [148][149]. Governance-focused reports further show that hospitals with formal AI governance structures are more than twice as likely to achieve 12-month ROI on AI investments than those relying on ad hoc oversight [150][151][152].

The extended architecture presented here with this research directly addresses the root causes of these ROI gaps. First, by centralising agent allocation, configuration, and monitoring under a unified orchestration runtime tied to the Compliance and Policy, and Governance layers of the baseline architecture designed by Dhar et al. (2026) [3][Fig], it removes the need for one-off integrations for each chatbot or model, reducing integration time and allowing governance teams to approve “patterns” rather than individual tools. Second, by encoding the risk tier, oversight requirement, and jurisdiction into the agent orchestration layer of the foundational architecture of Dhar et al. (2026) [3][Fig], makes it feasible to deploy higher-risk, higher-value Agentic capabilities such as triage decision support, imaging report summarisation, and claim reconciliation, in a controlled way that can pass internal risk committees and external conformity assessments, rather than being confined to low-impact administrative pilots. Third, by demonstrating the average time savings per task [Fig], especially in documentation-heavy and crucial coordination-intensive workflows (e.g., triage, imaging reports, discharge summaries, bed assignment, billing and claim submission), the architecture links agent patterns to measurable operational outcomes that CFOs and clinical leaders recognise as precursors to revenue uplift and cost savings.

Because the platform is designed to operate in on-premise, hybrid, and cloud-native deployments with confidential computing and MCP-based tool exposure, it is immediately applicable in jurisdictions with strict data localisation or cross-border transfer rules, expanding the addressable market beyond “cloud-permissive” settings and thereby increasing the expected ROI per design investment compared with more narrowly scoped cloud-only frameworks. In summary, the value proposition is not just better model metrics, but a shorter, safer path from experimental agent to globally compliant, production-grade Agentic workflows that deliver traceable time savings, improved throughput, and reduced rework across the hospital operations.

➢ *Safety, Confidentiality, and the “Where Can PHI Go?” Question*

A recurring concern in both policy papers and board-level discussions is the inability of many AI architectures to answer the question “*Where, exactly, could protected health information (PHI) go?*” in a precise, testable way [153][154]. Contemporary multi-agent frameworks often assume internet-reachable tools, shared state across agents, and opaque logging, which makes it difficult to reason about PHI flows in highly regulated environments [155][156]. Our architecture confronts this challenge by hard-wiring network, encryption, and key-management choices into the platform design:

- All PHI access passes through the Privacy-Preserving Data Fabric Layer [Fig][Fig] [Fig], which is deployed within private subnets and fronted by role and purpose-aware views, no agent whether vLLM-backed or tabular, can bypass this fabric.
- The Infrastructure and Runtime Layer [Fig] enforces encryption for data at-rest (AES-256-GCM with hardware-backed key management), encryption for data in-transit (TLS 1.3 with PQC-ready key exchange where available), and for data in-use (confidential VMs or TEEs for high-sensitivity workloads) [Fig] [Fig], ensuring that even privileged infrastructure operators have limited observability into PHI.
- MCP servers provide on-premise integration plane for tools and EHR APIs [Fig] [Fig], exposing only high-level tool manifests to the agent platform and never raw PHI to the public internet, in line with the principle that “*nothing that does not need the public internet should see the public internet at all*”.

By combining these controls with risk-tier-aware policy-as-code enforcement and exhaustive logging of every agent invocation, tool call, and policy decision, the extended architecture makes it realistic for hospital leaders to answer the PHI-flow question with specific, verifiable statements rather than general assurances. This is a differentiator relative to many recent multi-agent healthcare proposals, which focus on reasoning and collaboration but leave confidentiality and data-flow governance as future work.

➢ *Closing Statement*

In an environment where hospitals are inundated with AI proofs-of-concept and vendor frameworks that promise rapid automation but often stall at the governance gate, the extended architecture presented here offers a pragmatic, compliance-aligned pathway from isolated agent experiments to a governed Agentic ecosystem grounded in HIMS reality. By building directly on the seven-layered foundational architecture design [Fig] by Dhar et al. (2026) [3] and enriching it with an Agentic pattern catalogue, risk-stratification model, unified vLLM and MCP-enabled orchestration runtime, and implementation-grade code and datasets [Fig][Fig], this research provides both an intellectual contribution and a deployable blueprint. It demonstrates that it is possible to design an hospital AI infrastructure where performance, ROI, safety, and regulatory readiness reinforce rather than undermine one another, moving AI from experimental add-on to strategic, confidential intelligence infrastructure for mission-critical hospital information management systems.

## DISCLAIMER

➢ *Regulatory and Certification Obligations*

Deploying AI within clinical settings often necessitates regulatory scrutiny, certification, or formal approval based on the specific location, application, and risk level. While the proposed architecture aligned with compliance standards namely HIPAA, GDPR, EU AI Act, DPDP Act, DISHA India during the design and pilot deployment phase, this does not replace formal audits or mandatory regulatory submissions to authorities such as the FDA, EMA, or CDSCO. Healthcare institutions bear full responsibility for identifying applicable regulations and securing all requisite approvals, certifications, and Business Associate Agreements (BAAs) prior to live deployment.

➢ *Intellectual Property and Open Source*

Various platform components may be governed by copyrights, patents, or open-source agreements. Implementers are tasked with performing thorough intellectual property due diligence and obtaining all essential licenses and permissions prior to deployment. Furthermore, the architecture's dependence on commercial cloud infrastructures, EHR APIs, and large language models (LLMs) may present challenges related to licensing, data sovereignty, or vendor lock-in, requiring independent evaluation for each use case.

➢ *Funding and Financial Declaration*

This independent research received no funding, sponsorship, or commission from any external commercial, governmental, or institutional body. The findings, architectural designs, and analyses reflect the exclusive intellectual property of the authors and Instil-it, Hyderabad, India, and remain entirely free from financial or institutional bias. Any mention of particular technologies, vendors, or frameworks serves strictly academic purposes and does not imply affiliation or endorsement.

## AUTHOR CONTRIBUTIONS

Manideep Dhar served as the principal research conceptualizer, principal author, planner and research strategist for this study. He directed the problem analysis, statistical data analysis, architectural vision, module embedding, governance framework, compliance-oriented design principles, and also devised the overall AI structure, ultimately designing the multi-layered agentic AI framework for the hospital information management system (HIMS). Additionally, he formulated the privacy-preserving governance models, security architecture, and compliance-first and privacy-as-code fronted agentic strategies. He also contributed to the operationalisation of the architecture, including system integration planning, deployment-readiness assessment with alignment to the governance and regulations for real-world hospital environments.

Ritwik Singh acted as the chief architectural engineer, ensuring the model development lifecycle continuously adhered to technical, operational, and compliance standards. He supported the core AI system engineering and played a key role in the pilot deployment, actively monitoring system performance and addressing real-world challenges through iterative code optimization.

Sharat Chandra Kumar Manikonda contributed as the project owner and chief data scientist, overseeing all AI system development. He designed the integration architecture and interoperability strategies to guarantee seamless compatibility with legacy HIMS platforms. Furthermore, he spearheaded the technical implementation, model engineering workflows, orchestration feasibility, and pilot deployment governance, while significantly contributing to the operationalisation of the architecture, including system integration planning, workflow optimisation logic, and deployment-readiness assessment for real-world hospital environments.

All authors jointly participated in the research, design, validation, and refinement of the proposed architecture. They synthesized their respective domain expertise to ensure the study exhibited technical rigor, operational feasibility, and strict adherence to governance standards.